%% file: colm2026_conference.tex
\documentclass{article}
\PassOptionsToPackage{table}{xcolor}
\usepackage[preprint]{colm2026_conference}

\usepackage{microtype}
\usepackage{graphicx}
\usepackage{amsmath}
\usepackage{amssymb}
\usepackage{pifont}
\newcommand{\cmark}{\ding{51}}
\newcommand{\xmark}{\ding{55}}
\usepackage{enumitem}
\usepackage{booktabs}
\usepackage{amsthm}
\usepackage{makecell}

\definecolor{mydarkblue}{rgb}{0,0.08,0.45}
\usepackage[colorlinks=true,linkcolor=mydarkblue,citecolor=mydarkblue,filecolor=mydarkblue,urlcolor=mydarkblue]{hyperref}

\usepackage{url}
\usepackage{tabularx}
\usepackage{multirow}
\usepackage{color-edits}
\usepackage{framed}
\usepackage{tikz}
\usetikzlibrary{calc, positioning, shadows.blur, shapes.geometric, arrows.meta}
\usepackage{fontawesome5}

\newcommand{\circled}[1]{\ding{\the\numexpr#1+191}}
\newcommand{\hflogo}{\raisebox{-0.2ex}{\includegraphics[height=1em]{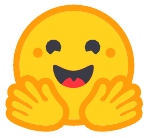}}}

\input{pre}

\addauthor{sw}{purple}
\addauthor{xz}{blue}
\addauthor{cm}{cyan}
\addauthor{gn}{magenta}

\newenvironment{itemize*}%
 {\leftmargini=10pt\begin{itemize}%
  \setlength{\itemsep}{0pt}%
  \setlength{\parskip}{0pt}%
  }%
 {\end{itemize}}
\newenvironment{enumerate*}%
 {\begin{enumerate}%
  \setlength{\itemsep}{0pt}%
  \setlength{\parskip}{0pt}}%
 {\end{enumerate}}

\begin{document}

\title{Reinforcing Human Behavior Simulation via Verbal Feedback}

\author{\textbf{Weiwei Sun}$^{1}$\thanks{Equal contribution.} \quad
\textbf{Xuhui Zhou}$^{1}$\footnotemark[1] \\
\textbf{Jiarui Liu}$^{1}$ \quad
\textbf{Weihua Du}$^{1}$ \quad
\textbf{Haojia Sun}$^{1}$ \quad
\textbf{Yiqing Xie}$^{1}$ \quad
\textbf{Qianou Ma}$^{1}$ \\
\textbf{Sihao Chen}$^{2}$ \quad
\textbf{Mengting Wan}$^{2}$ \quad
\textbf{Longqi Yang}$^{2}$ \quad
\textbf{Pei Zhou}$^{2}$ \\
\textbf{Sherry Wu}$^{1}$ \quad
\textbf{Sean Welleck}$^{1}$ \quad
\textbf{Graham Neubig}$^{1}$ \quad
\textbf{Yiming Yang}$^{1}$ \quad
\textbf{Maarten Sap}$^{1}$ \\[3pt]
$^{1}$Carnegie Mellon University \quad $^{2}$Microsoft \\[3pt]
\texttt{\{weiweis,xuhuiz\}@andrew.cmu.edu}\\[3pt]
\faGithub~ \href{https://github.com/sunnweiwei/OdysSim}{\texttt{Code}} \quad
\hflogo~ \href{https://huggingface.co/sunweiwei/Ditto-8B}{\texttt{Model}} \quad
\hflogo~ \href{https://huggingface.co/sunweiwei/Soul}{\texttt{Data}}
}
\maketitle

\input{sections/abstract}
\input{sections/introduction}
\input{sections/related_work}
\input{sections/method}

\input{sections/evaluation_benchmarks}
\input{sections/experimental_setup}
\input{sections/results}

% \input{sections/analysis}
\input{sections/conclusion}

\bibliographystyle{colm2026_conference}
\bibliography{colm2026_conference}

\input{sections/appendix}

\end{document}

%% file: pre.tex
\definecolor{gred}{RGB}{250, 210, 207}
\definecolor{coolblue1}{rgb}{0.91, 0.94, 0.98}
\definecolor{coolblue2}{rgb}{0.76, 0.85, 0.94}
\definecolor{coolblue3}{rgb}{0.54, 0.72, 0.87}
\definecolor{coolblue4}{rgb}{1, 1, 1}

\usepackage{xspace}
\usepackage{cleveref}

%% file: sections/abstract.tex
% ===========================================================
% Abstract
% ===========================================================

\begin{abstract}
Humans learn social norms and behaviors from verbal feedback (e.g., a parent saying ``that was rude'' or a friend explaining ``here's why that hurt''). 
Yet, learning from feedback for LLMs has largely focused on domains like code and math where RL rewards are directly verifiable and condensed into scalar values.
As LLMs are increasingly also used for simulating human behavior, e.g., standing in for users, patients, students, and other personas, there is a pressing need to make them more human-like, which requires embracing a fundamentally different kind of signal: feedback that is verbal, subjective, and multi-faceted.
We present \textbf{\textsc{Ditto}}, a model trained by treating verbal feedback as a first-class signal in reinforcement learning. After each rollout, \textsc{Ditto} receives verbal feedback and generates a feedback-conditioned improved rollout; both outputs are jointly optimized with GRPO, distilling verbal guidance into the base policy without requiring feedback at test time. We also introduce \textbf{\textsc{Soul}} (Simulation gym Of hUman-Like behavior), a unified benchmark and training data suite spanning \textit{10 tasks} across six categories: Theory of Mind, character role play, social skill, learner simulation, user simulation, and persona simulation. \textsc{Ditto} achieves an average 36\% improvement over the base model and exceeds GPT-5.4 on 6 of 10 \textsc{Soul} benchmarks, demonstrating that RL with verbal feedback is a promising direction for training LLMs to simulate human behavior.
\end{abstract}

%% file: sections/introduction.tex
% ===========================================================
% Introduction
% ===========================================================

\section{Introduction}
\label{sec:introduction}

% Maarten's feedback on intro flow:
% I like this current flow, but I feel like it focuses more on the application ie building better simulators, than the method? This may be fine, but there are other flows:
% - SOTW: RL typically uses verifiable rewards
% - Challenge: as we need more human-like LLMs, we need to mirror how humans learn
% another option (more cognitive? more aligned with my suggested abstract flow):
% - SOTW: humans learn via complex verbal feedback, but LLMs learn from verifiable rewards (RL background)
% - Challenge: now that we need more user simulators, we need to bridge this gap to have methods for how humans learn for LLMs.

Modeling human behavior has long been recognized as a crucial step toward imbuing artificial intelligence with social intelligence \citep{tomasello2019becoming, nass2000machines}.
From the Turing test to modern user simulators, the drive to faithfully model how people think, speak, and act has evolved into a practical necessity for training and evaluating collaborative AI systems.
Today, LLMs are increasingly called upon to serve as these vital proxies---standing in for users in dialogue systems \citep{davidson2023user, sekulic2024reliable}, for patients in clinical settings \citep{schmidgall2024agentclinic}, for students in educational platforms \citep{tack2024mistakes, dou2025simulatorarena}, and for diverse personas in social and safety testing \citep{zhou2024sotopia, zhou2025haicosystem, Wang2023UserBS}.

% Simulating human behavior has been a long-standing aspiration of artificial intelligence \citep{tomasello2019becoming, nass2000machines}.
% From the Turing test to modern user simulators, the question of whether machines can faithfully replicate how people think, speak, and act remains one of the field's fundamental challenges.
% Today, LLMs are increasingly called upon to do exactly this---standing in for users in dialogue systems \citep{davidson2023user, sekulic2024reliable}, for patients in clinical settings \citep{schmidgall2024agentclinic}, for students in educational platforms \citep{tack2024mistakes, dou2025simulatorarena}, and for diverse personas in social and safety testing \citep{zhou2024sotopia, zhou2025haicosystem, Wang2023UserBS}.

Yet current LLM simulators fall short, exhibiting systematic Sim2Real gaps---from homogeneous outputs lacking individual variation \citep{jiang2025hivemind}, to misaligned cognitive biases \citep{tjuatja2024llms}, superhuman behaviors \citep{oh2025superhuman}, and consistent divergences from real users in downstream tasks \citep{zhou2026mindsim2realgapuser}.
One reason for this gap may lie in \emph{how} these models are trained.
Reinforcement learning for LLMs, whether RLHF \citep{ouyang2022rlhf}, DPO \citep{rafailov2023dpo}, or RL with verifiable rewards \citep{deepseek_r1, shao2024deepseekmath}, reduces all feedback to a scalar reward or binary preference.
This suffices when correctness is verifiable: a proof is right or wrong, a test passes or fails.
But simulating human behavior is not about solving a puzzle correctly; it is about acting in ways that are socially grounded, individually varied, and contextually appropriate.
A score of 0.6 conveys almost nothing about \emph{why} a simulated user was too polite or too predictable, let alone how to fix it.
Humans themselves learn social behavior not from scores but from verbal feedback, through explanations, corrections, and dialogue \citep{tomasello2019becoming}, and we argue that training human simulators should follow a similar principle.

\begin{figure*}[t!]
\centering
% \vspace{-2em}
\includegraphics[width=\textwidth]{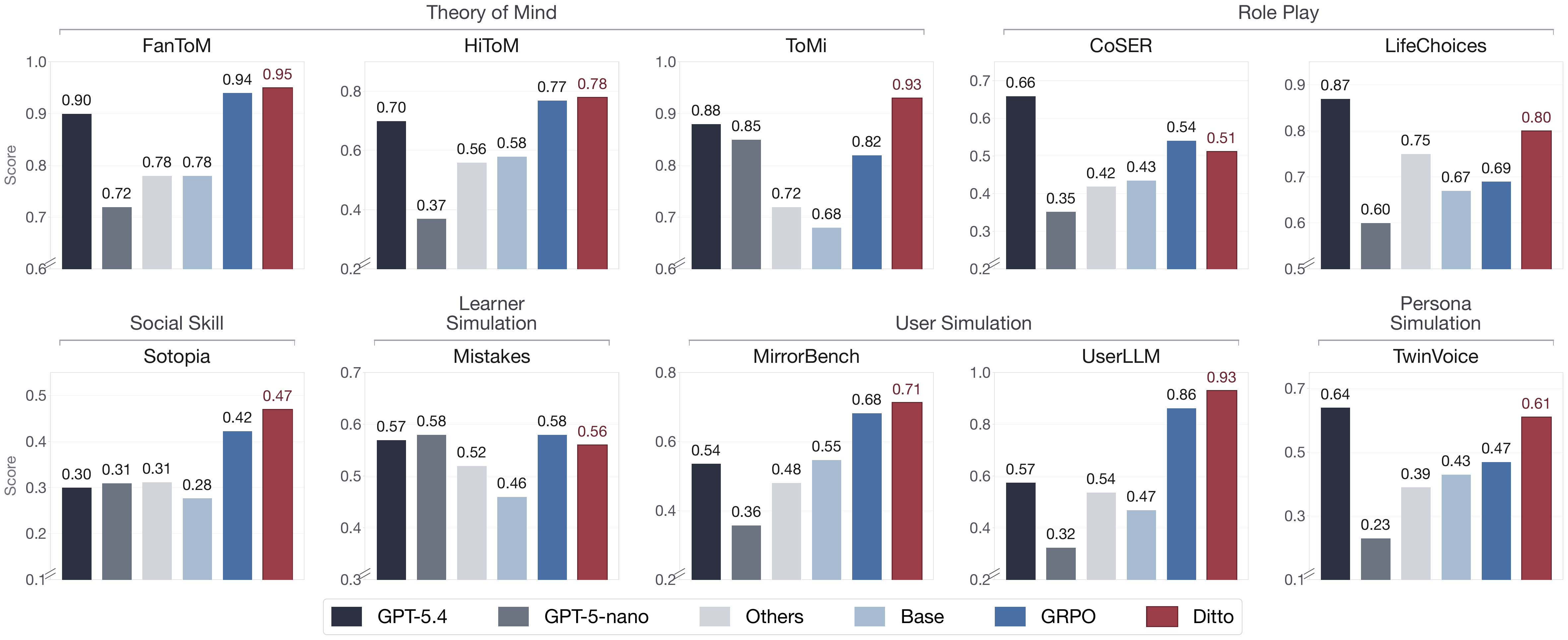}
% \vspace{-2em}
\caption{Benchmark results on human simulation tasks. Ditto wins 6 of 10 benchmarks, surpassing both frontier closed-source models and specialized human-simulation baselines.}\label{fig:benchmark}
% \vspace{-0.5em}
\end{figure*}

As shown in \Cref{fig:method_overview}, we present \textbf{\textsc{Ditto}}, a model that incorporates \textbf{verbal feedback} as a first-class signal in reinforcement learning for human behavior simulation \citep{song2026rltf, shi2026experiential}.
After each rollout, \textsc{Ditto} receives verbal feedback and generates an improved rollout conditioned on it; both outputs are jointly optimized via GRPO, so that the base policy internalizes the verbal guidance without requiring it at test time (\S\ref{sec:method}).
Crucially, our approach is flexible across task formats: it handles discriminative tasks (multiple-choice, binary classification), generative tasks (open-ended dialogue), and multi-turn interactions where the model must sustain human-like behavior across an entire conversation, all within a single unified training framework.

To train and evaluate \textsc{Ditto}, we introduce \textbf{\textsc{Soul}} (\textbf{S}imulation gym \textbf{O}f h\textbf{U}man-\textbf{L}ike behavior; \Cref{fig:benchmark_overview}), a unified benchmark and training data suite comprising \textit{10 tasks} across six categories of 
% human simulation
human behavior modeling
: theory of mind \citep{kim2023fantom, street2024hitom, jin2024mmtom, nematzadeh2018tomi, sclar2023paratomi}, character role play \citep{wang2025coser}, social skill \citep{zhou2024sotopia}, learner simulation \citep{tack2024mistakes}, user simulation \citep{hathidara2025mirrorbench, naous2025flipping}, and persona simulation \citep{chen2025twinvoice}.
A key bottleneck for RL-based human simulation has been the absence of training data: most existing human simulation benchmarks provide only evaluation sets, not training splits.
\textsc{Soul} addresses this by pairing every task with curated RL training data, adapting official splits where available and constructing new ones where not, such as 6,000 synthetic belief-tracking scenarios for HiToM, and 1,358 persona items for TwinVoice built from three external corpora via a two-stage LLM generation and filtering pipeline.
The resulting mixture spans discriminative, generative, multi-turn, and multi-agent formats within a single unified RL training framework (\S\ref{sec:soul}).

Our experiments on \textsc{Soul} yield several findings.
%
% First, \textsc{Ditto} achieves an average score of 0.726, a 36\% improvement over the base model, while matching or exceeding GPT-5.4 on 6 of 10 tasks.
% Gains are broad: \textsc{Ditto} outperforms GPT-5.4 on all three Theory of Mind tasks, on both user simulation tasks (e.g., 0.930 vs.\ 0.575 on UserLLM), and on Sotopia social skill (0.470 vs.\ 0.300), spanning discriminative, generative, and multi-turn formats.
%
First, our Qwen3-8B \textsc{Ditto} model achieves an average normalized score of 0.726, a 36\% improvement over the base model, while matching or exceeding GPT-5.4 on 6 of 10 tasks. Gains are broad: \textsc{Ditto} outperforms GPT-5.4 on all three Theory of Mind tasks, on both user simulation tasks (e.g., scoring 0.930 compared to GPT-5.4's 0.575 on UserLLM), and on Sotopia social skill (0.470 vs.\ 0.300), spanning discriminative, generative, and multi-turn formats.
Second, verbal feedback consistently improves over standard GRPO, with the largest gains on generative and multi-turn tasks such as ToMi (+11 points), LifeChoices (+11), and TwinVoice (+14).
Third, analysis of training dynamics shows that verbal feedback accelerates learning---on Sotopia, \textsc{Ditto} outperforms GRPO from the earliest training steps---and helps mitigate safety risks: \textsc{Ditto} better preserves secret-keeping throughout RL optimization, whereas standard GRPO degrades on this dimension.

\paragraph{Contributions.}
\begin{enumerate}[label=(\roman*)]
    \item We show that \textbf{verbal feedback is a crucial training signal} for human behavior simulation, providing actionable, dimension-specific guidance beyond scalar rewards, improving learning speed and reducing safety risks such as secret leakage.
    \item We introduce \textbf{\textsc{Ditto}}, which distills verbal feedback into the base policy via feedback-conditioned rollouts jointly optimized with GRPO, achieving a 36\% improvement over the base model and matching or exceeding GPT-5.4 on 6/10 tasks with an 8B model.
    \item We release \textbf{\textsc{Soul}}, a unified training and evaluation environment for human-like behavior across 10 tasks in six categories, covering discriminative, generative, multi-turn, and multi-agent settings, with curated RL training data for each task.
\end{enumerate}

%% file: sections/related_work.tex
% ===========================================================
% Related Work
% ===========================================================

\section{Related Work}
\label{sec:related_work}

\subsection{Incorporating Verbal Feedback in RL Training}
Standard RL for language models---whether RLHF \citep{ouyang2022rlhf}, DPO \citep{rafailov2023dpo}, or GRPO \citep{deepseek_r1, shao2024deepseekmath}---reduces all feedback to scalar rewards or binary preferences.
To handle complex feedback signals, another line of work decomposes the scalar reward into finer-grained or multi-dimensional signals while remaining in the scalar regime \citep{wu2023finegrained, yu2025sotopiarl, sun2025ppp}.
These methods enrich the reward structure but still compress each dimension into a scalar, losing rich nature of verbal feedback.

To incorporate verbal feedback into improving LLM agents, prior work has explored supervised fine-tuning \citep{scheurer2023training, liu2023chain} or inference-time self-reflection \citep{shinn2023reflexion, madaan2023selfrefine}, but these approaches lack the exploration capacity of RL or do not update model weights from experience. 
A recent line of work instead incorporates verbal feedback directly into RL optimization. 
Self-Rewarding Language Models \citep{yuan2024selfrewarding} generate verbal self-judgments to construct preference pairs for iterative DPO.
Experiential RL \citep{shi2026experiential} introduces an experience-reflection-consolidation loop mixed RL and supervised fine-tuning.
Concurrent to our approach, RLTF \citep{song2026rltf} formalizes reinforcement learning from text feedback and proposes self-distillation (RLTF-SD), which trains the single-turn policy to match its own feedback-conditioned second-turn generations.
However, these works focus extensively on domains with verifiable rewards, such as mathematics and reasoning puzzles.
Human behavior simulation poses fundamentally different challenges: interactions are multi-turn, rewards are subjective and multi-dimensional.

\subsection{Simulating Human Behavior with LLMs}

LLM-based human simulation spans a broad landscape of tasks.
User simulation has evolved from rule-based and statistical approaches for dialogue management \citep{schatzmann2006survey, li2016user} to LLM-based simulators for task-oriented dialogue \citep{davidson2023user, sekulic2024reliable,Sun2022MetaphoricalUS}, information-seeking \citep{Sun2021SimulatingUS, Zhang2024USimAgentLL, Ren2024BASESLW}, and recommendation \citep{Wang2023UserBS, Bougie2025SimUSERSU}.
Beyond user simulation, LLMs are increasingly used for social interaction \citep{zhou2024sotopia, zhou2025haicosystem}, character role-play \citep{wang2025coser}, persona simulation \citep{chen2024persona}, and learner simulation \citep{tack2024mistakes}.

However, recent efforts have revealed systematic \emph{Sim2Real gaps}: LLM-simulated human behavior shows systematic differences compared to real humans \citep{seshadri2026lost, tjuatja2024llms, zhou2026mindsim2realgapuser}, which undermine the utility of simulators for downstream evaluation and training.
Training-based approaches have begun to address this---UserLLM \citep{naous2025flipping} via SFT on human conversations and Sotopia-RL \citep{yu2025sotopiarl} via RL with multi-dimensional rewards. However, each targets a narrow set of tasks (e.g., \citet{yu2025sotopiarl} focus only on social skills, \citet{naous2025flipping} focus only on user simulation).
Our work incorporates verbal feedback into RL to improve a single model across the 10 diverse tasks in \textsc{Soul}, spanning six categories of human behavior modeling.

%% file: sections/method.tex
% ===========================================================
% Method
% ===========================================================
\input{figures/method_overview}

\section{Method}
\label{sec:method}

\subsection{Problem Formulation}

We study the problem of training a language model for human simulation tasks. Given an input $x$ (e.g., an instruction, scenario, dialogue history, or user profile), the model generates an output trajectory
$$
y = (y_1, \dots, y_T) \sim \pi_\theta(\cdot \mid x),
$$
where $\pi_\theta$ is the current policy.

Each sampled trajectory is evaluated by a task-specific judge or simulator, which returns supervision for learning. In the standard RL setting, this supervision is reduced to a scalar reward
$$
r(x,y) \in \mathbb{R}.
$$
Our goal is to train the policy $\pi_\theta$ to maximize task performance in these human simulation tasks, while allowing training to exploit richer judge supervision when available.

\subsection{Verbal Feedback}

In many human simulation tasks, the judge provides not only a scalar reward but also rich verbal feedback, such as dimension-wise scores, textual critiques, and task-specific suggestions. Formally, for a rollout $y \sim \pi_\theta(\cdot \mid x)$, the judge returns
\begin{equation}
(r, h) = \mathcal{J}(x, y),
\end{equation}
where $r$ is a scalar reward and $h$ is structured feedback describing what was done well, what went wrong, and how the behavior can be improved.

This feedback is often highly informative. For example, in human behavior simulation, it may reflect dimensions such as believability, relationship maintenance, knowledge consistency, goal achievement, and adherence to social norms. Standard RL methods typically compress all such information into a single scalar reward, which can lose useful learning signal.

To ensure robustness of learning, we cast $h$ as privileged information~\citep{vapnik2009new,Vapnik2015LearningUP}: it is available during training but not at test time. Our goal is therefore to use this verbal feedback to improve the base policy $\pi_\theta(\cdot \mid x)$, while keeping inference-time generation unchanged.

% For each sampled rollout $y_0 \sim \pi_\theta(\cdot \mid x)$, a task-specific LLM judge evaluates the output and provides:
% \begin{enumerate*}
% \item a scalar reward $r_0 = R(x, y_0)$, and
% \item structured verbal feedback $c$, which may include dimension-wise scores, textual critiques, and task-specific references.
% \end{enumerate*}
% This feedback serves as actionable guidance describing what was done well, what went wrong, and how the behavior should be improved, and the policy can condition on it directly during learning.

% The reflection serves as \emph{privileged information} \citep{vapnik2009new,Vapnik2015LearningUP,Shenfeld2026SelfDistillationEC,Ye2026OnPolicyCD,Hubotter2026ReinforcementLV}: it is available during training but not at test time. Our goal is to leverage this privileged signal to improve the base policy $\pi_\theta(\cdot \mid x)$ so that it performs well without access to $h$ at inference.

\subsection{Learning from Verbal Feedback}

% Our goal is to optimize the base policy $\pi_\theta(y \mid x)$ using both reward and feedback during training, while keeping the inference-time policy unchanged. That is, we want the policy to learn from feedback and internalize the resulting improvements, without requiring feedback as input at test time.
We aim to optimize the base policy $\pi_\theta(y \mid x)$ using reward and feedback during training, so the policy internalizes feedback-driven improvements without requiring feedback at inference time.

To achieve this, for each sampled rollout
\begin{equation}
y_0 \sim \pi_\theta(\cdot \mid x),
\end{equation}
we ask the judge to provide reward and feedback
\begin{equation}
(r_0, h) = \mathcal{J}(x, y_0),
\end{equation}
and then sample a second rollout conditioned on the same feedback:
\begin{equation}
y_1 \sim \pi_\theta(\cdot \mid x, h),
\qquad
r_1 = R(x, y_1).
\end{equation}
% Here, $y_0$ is the original rollout from the base policy, while $y_1$ is a feedback-conditioned rollout that reflects how the model responds when given explicit guidance for improvement.
% Here, $y_0$ is the original rollout from the base policy, which we refer to as the \emph{student} rollout, while $y_1$ is the feedback-conditioned rollout, which we refer to as the \emph{teacher} rollout. The teacher rollout reflects how the same policy responds when given explicit verbal guidance for improvement.
Here, $y_0$ is the original rollout from the base policy, which we refer to as the \emph{draft} rollout, while $y_1$ is the feedback-conditioned rollout, which we refer to as the \emph{refined} rollout. Both rollouts are generated by the same policy with shared model weights; they differ only in their prompting context, where the refined rollout additionally uses explicit verbal feedback to guide improvement.

We optimize both behaviors during training. First, we train the base policy to learn from feedback-induced improvements without conditioning on feedback at inference. Second, we include an additional GRPO objective on feedback-conditioned rollouts themselves, so the model also becomes better at generating improved responses from feedback during training.

Concretely, for each prompt $x$, we first sample $G$ original rollouts $\{y_{i,0}\}_{i=1}^G$ from the base policy. For each $y_{i,0}$, we obtain judge feedback $h_i$ and generate a corresponding feedback-conditioned rollout $y_{i,1}$. This gives the training group
\begin{equation}
\mathcal{G}(x) = \{y_{i,0}, y_{i,1}\}_{i=1}^{G},
\end{equation}
which contains both the original rollouts and their feedback-conditioned counterparts.

We compute group-relative advantages over samples in $\mathcal{G}(x)$:
We compute group-relative advantages over all samples in $\mathcal{G}(x)$:
\begin{equation}
A_{i,j}
=
\frac{r_{i,j} - \mu_r}{\sigma_r},
\qquad
\mu_r
=
\frac{1}{2G}
\sum_{i=1}^{G}
\sum_{j=0}^{1}
r_{i,j},
\qquad
\sigma_r
=
\mathrm{std}(\{r_{i,j}\}_{i=1,j=0}^{G,1}).
\end{equation}

We then apply the standard clipped GRPO objective to this group:
\begin{equation}
\mathcal{L}_{\mathrm{group}}
=
-\mathbb{E}_x
\left[
\sum_{i=1}^{G}
\sum_{j \in \{0,1\}}
\sum_{t=1}^{|y_{i,j}|}
\min\!\Big(
\rho_{i,j,t} A_{i,j},
\mathrm{clip}(\rho_{i,j,t}, 1-\epsilon, 1+\epsilon) A_{i,j}
\Big)
\right],
\end{equation}
where
\begin{equation}
\rho_{i,j,t}
=
\frac{\pi_\theta(y_{i,j,t} \mid x, y_{i,j,<t})}
{\pi_{\mathrm{old}}(y_{i,j,t} \mid x, y_{i,j,<t})}.
\end{equation}

In addition, we apply an extra GRPO update on the feedback-conditioned rollouts $\{y_{i,1}\}_{i=1}^G$ alone, treating them as a separate GRPO group:
\begin{equation}
\mathcal{L}_{\mathrm{fb}}
=
-\mathbb{E}_x
\left[
\sum_{i=1}^{G}
\sum_{t=1}^{|y_{i,1}|}
\min\!\Big(
\hat{\rho}_{i,t} \hat{A}_{i,1},
\mathrm{clip}(\hat{\rho}_{i,t}, 1-\epsilon, 1+\epsilon)\hat{A}_{i,1}
\Big)
\right],
\end{equation}
where $\hat{A}_{i,1}$ is computed by normalizing rewards within the feedback-conditioned group $\{r_{i,1}\}_{i=1}^G$. The final training loss is
\begin{equation}
\mathcal{L}
=
\mathcal{L}_{\mathrm{group}}
+ \mathcal{L}_{\mathrm{fb}}.
\end{equation}

In this way, feedback serves two roles during training: it produces improved rollouts that help the base policy learn better behavior, and it also supports a direct RL signal for improving the model's ability to respond to feedback itself.

\subsection{Connections to Other RL Approaches} %Discussion}
Our approach is inspired by, or developed concurrently with, several recent works on RL with textual feedback. We discuss the most relevant connections below.

\paragraph{Connection to RLTF}
Our method is related to RL from Text Feedback (RLTF; \citealp{song2026rltf}), which also uses training-time text feedback to improve single-turn performance. Both methods first generate a student output, then a feedback-conditioned teacher output, and train the policy to absorb the teacher’s improvement. Our method differs in three ways: (1) we use GRPO over the joint group ${y_0, y_1}$ rather than AWR (advantage-weighted regression, \citep{Peng2019AdvantageWeightedRS}); (2) our teacher conditions on a judge-generated reflection $h$ rather than raw feedback; and (3) unlike RLTF-SD, which targets tasks with \textit{verifiable rewards}, we focus on social simulation, where feedback is noisy, subjective, and multi-dimensional.
We compare against an RLTF-SD variant in our ablation study (\cref{results:ablation}).

\paragraph{Connection to ERL}
Experiential RL (ERL)~\citep{shi2026experiential} instead uses supervised fine-tuning (SFT) on reject-sampled teacher rollouts to internalize feedback into the student model. In contrast, we train on all teacher rollouts with an RL objective, allowing us to differentiate teacher rollout quality in continuous-reward settings and improve sample efficiency. We also compare against an ERL variant in our ablation study.

\paragraph{Connection to SDPO}
Several works propose self-distillation policy optimization, distilling textual feedback into the student with a reverse KL objective~\citep{vapnik2009new,Vapnik2015LearningUP,Shenfeld2026SelfDistillationEC,Ye2026OnPolicyCD,Hubotter2026ReinforcementLV}. In contrast, our approach uses a forward alternative: sampling from the teacher rather than the student, and optimizing with group-normalized advantage. In addition, we optimize the feedback-conditioned model via $\mathcal{L}_{\mathrm{fb}}$ rather than using an EMA model. We also compare against SDPO-like variants in our ablation study.

%% file: figures/method_overview.tex
\begin{figure*}[t]
\centering
\resizebox{0.95\textwidth}{!}{%
\begin{tikzpicture}[
  every node/.style={font=\sffamily},
  card/.style={
    rounded corners=5pt,
    minimum height=0.8cm,
    font=\sffamily\footnotesize,
    inner sep=5pt,
    drop shadow={shadow xshift=0.3pt, shadow yshift=-0.3pt, opacity=0.07},
    line width=0.5pt,
  },
  pill/.style={
    rounded corners=2pt,
    font=\sffamily\tiny\bfseries,
    inner sep=2pt,
    text=white,
  },
  arr/.style={-{Stealth[length=4pt, width=3pt]}, line width=0.8pt, rounded corners=2pt},
  lbl/.style={font=\sffamily\scriptsize, text=black!50},
]

% ===== Colors =====
\definecolor{polcyan}{RGB}{16,185,185}
\definecolor{polcyanL}{RGB}{232,253,253}
\definecolor{judgamb}{RGB}{245,158,11}
\definecolor{judgambL}{RGB}{255,244,224}
\definecolor{fborg}{RGB}{234,88,12}
\definecolor{fborgL}{RGB}{255,237,213}
\definecolor{grpopurp}{RGB}{124,58,237}
\definecolor{grpopurpL}{RGB}{243,238,255}
\definecolor{soulblue}{RGB}{37,99,235}
\definecolor{soulblueL}{RGB}{228,237,255}
\definecolor{studblue}{RGB}{59,130,246}
\definecolor{teachrose}{RGB}{225,29,72}
\definecolor{testgreen}{RGB}{22,163,74}
\definecolor{testgreenL}{RGB}{225,252,235}
\definecolor{bgcard}{RGB}{250,250,252}

% ===== Layout positions =====
\def\rowA{1.6}     % Student path
\def\rowB{0.15}    % Verbal feedback center
\def\rowC{-1.3}    % Teacher path

\def\colPol{3.2}
\def\colOut{5.2}
\def\colR{7.5}     % right column: GRPO + Test

% =========================================================
% LEFT: SOUL TASKS
% =========================================================
\node[card, fill=soulblueL, draw=soulblue!30,
      minimum width=2.2cm, minimum height=4.4cm]
      (soul) at (0, 0.0) {};
\node[font=\bfseries\small, text=soulblue] at (0, 1.75) {\textsc{Soul}};
\node[pill, fill=soulblue!75] at (0, 1.25) {\faIcon{brain}~Theory of Mind};
\node[pill, fill=soulblue!65] at (0, 0.78) {\faIcon{theater-masks}~Role Play};
\node[pill, fill=soulblue!55] at (0, 0.31) {\faIcon{handshake}~Social Skill};
\node[pill, fill=soulblue!65] at (0, -0.16) {\faIcon{graduation-cap}~Learner Sim};
\node[pill, fill=soulblue!75] at (0, -0.63) {\faIcon{user}~User Sim};
\node[pill, fill=soulblue!65] at (0, -1.10) {\faIcon{id-badge}~Persona Sim};
\node[font=\sffamily\tiny, text=black!30] at (0, -1.55) {10 tasks};

% ===== Ditto mascot =====

% =========================================================
% RIGHT COLUMN: GRPO (top half) + TEST (bottom half)
% =========================================================
% GRPO box — right side, top portion matching SOUL card
\node[card, fill=grpopurpL, draw=grpopurp!35,
      minimum width=2.4cm, minimum height=2.0cm,
      text width=2.0cm, align=center,
      anchor=north]
      (grpo) at (\colR, 2.2) {};

\node[font=\sffamily\footnotesize\bfseries, text=grpopurp, anchor=north]
      at ($(grpo.north)+(0,-0.25)$) {Joint GRPO};
\node[font=\sffamily\tiny, text=black!50, text width=2.0cm, align=center, anchor=north]
      at ($(grpo.north)+(0,-0.7)$) {
  group-relative\\optimization over\\$\{(y_0, r_0),\;(y_1, r_1)\}$
};

% =========================================================
% ROW A: STUDENT PATH
% =========================================================
\def\branchX{1.8}
\coordinate (branch) at (\branchX, \rowA);

% SOUL → branch
\draw[soulblue!45, line width=0.8pt] (1.25, \rowA) -- (branch);
\node[circle, fill=soulblue!50, inner sep=0pt, minimum size=4pt] at (branch) {};

% Policy
\node[card, fill=polcyanL, draw=polcyan!40,
      minimum width=1.4cm, align=center]
      (pol1) at (\colPol, \rowA) {\textbf{$\pi_\theta$}};

% branch → pol1
\draw[arr, color=soulblue!45] (branch) -- node[above, lbl] {$x$} (pol1.west);

% y0
\node[card, fill=bgcard, draw=studblue!35, minimum width=1.0cm]
      (y0) at (\colOut, \rowA) {$y_0$};
\node[pill, fill=studblue, anchor=south] at ($(y0.north)+(0,0.04)$) {\scriptsize student};

\draw[arr, color=studblue!55] (pol1.east) -- (y0.west);

% r0 arrow going right into GRPO
\draw[arr, color=judgamb!50] (y0.east) -- node[above, lbl] {$r_0$} (grpo.west |- y0);

% =========================================================
% ROW B: VERBAL FEEDBACK (center)
% =========================================================
\node[card, fill=fborgL, draw=fborg!35,
      minimum width=3.2cm, minimum height=1.3cm]
      (fb) at (4.2, \rowB) {};

\node[font=\sffamily\scriptsize\bfseries, text=fborg, anchor=west]
      at ($(fb.north west)+(0.1,-0.18)$)
      {\faIcon{comment-dots}~Verbal Feedback};

\node[font=\sffamily\tiny, text=black!55, text width=2.9cm, align=left, anchor=north west]
      at ($(fb.north west)+(0.12,-0.45)$) {
  $h$: \textit{``push back more on conflicts\ldots''}
};
\node[pill, fill=fborg!60, anchor=south east]
      at ($(fb.south east)+(-0.08, 0.08)$)
      {\tiny LLM-as-Judge};

% y0 → feedback
\draw[arr, color=fborg!50] (y0.south) -- ($(fb.north)+(1.0, 0)$);

% =========================================================
% ROW C: TEACHER PATH
% =========================================================
% x branches down
\coordinate (branchC) at (\branchX, \rowC);
\draw[soulblue!45, line width=0.8pt] (branch) -- (branchC);

% h from feedback left side, corner down to merge
\draw[fborg!50, line width=0.8pt, rounded corners=3pt]
  (fb.west) -- (\branchX, \rowB) -- (branchC);
\node[lbl, anchor=east] at ({\branchX-0.1}, {(\rowB+\rowC)/2}) {$h$};

% merge dot
\node[circle, fill=fborg!50, inner sep=0pt, minimum size=4pt] at (branchC) {};
\node[font=\sffamily\scriptsize\bfseries, text=fborg!70, anchor=south]
      at ($(branchC)+(0.35, 0.12)$) {$x\!+\!h$};

% Policy
\node[card, fill=polcyanL, draw=polcyan!40,
      minimum width=1.4cm, align=center]
      (pol2) at (\colPol, \rowC) {\textbf{$\pi_\theta$}};

\draw[arr, color=fborg!50] (branchC) -- (pol2.west);

% y1
\node[card, fill=bgcard, draw=teachrose!35, minimum width=1.0cm]
      (y1) at (\colOut, \rowC) {$y_1$};
\node[pill, fill=teachrose, anchor=north] at ($(y1.south)+(0,-0.04)$) {\scriptsize teacher};

\draw[arr, color=teachrose!55] (pol2.east) -- (y1.west);

% r1 arrow — right then up into GRPO left side
\draw[arr, color=judgamb!50, rounded corners=4pt]
  (y1.east) -- ++(0.3, 0) node[above, lbl] {$r_1$} |- ($(grpo.south west)!0.15!(grpo.north west)$);

% =========================================================
% TEST TIME — directly below GRPO
% =========================================================
\node[card, fill=testgreenL, draw=testgreen!30, dashed,
      minimum width=2.4cm, align=center,
      anchor=south]
      (test) at (\colR, -2.2) {
  \faIcon{bolt}~\textbf{Test time}\\[2pt]
  {\scriptsize $x \to \pi_\theta \to y$}\\[1pt]
  {\tiny\color{black!40}(no feedback needed)}
};
% ===== Ditto mascot beside the GRPO→Test arrow =====
\draw[arr, color=grpopurp!40] (grpo.south) -- (test.north);
\node[anchor=west, inner sep=0pt] (dittoimg) at ($(grpo.south)!0.5!(test.north)+(0.15,0)$)
  {\includegraphics[height=0.6cm]{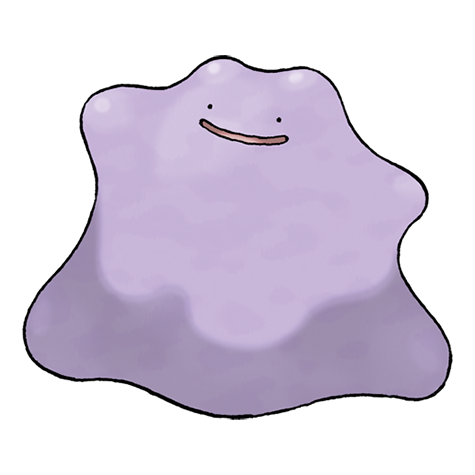}};
\node[font=\sffamily\tiny\bfseries, text=grpopurp, anchor=north] at ($(dittoimg.south)+(0,0.18)$) 
  {\textsc{Ditto}};

\end{tikzpicture}%
}% end resizebox
\caption{Overview of \textsc{Ditto}. \circled{1}~Given prompt $x$ from \textsc{Soul}, the policy generates a \textcolor[RGB]{59,130,246}{student} rollout $y_0$, scored as $r_0$. \circled{2}~An LLM judge produces \textbf{verbal feedback}: per-dimension critiques and a reflection $h$. \circled{3}~$h$ is concatenated with $x$ and fed into $\pi_\theta$ to generate a \textcolor[RGB]{225,29,72}{teacher} rollout $y_1$, scored as $r_1$. \circled{4}~Both are jointly optimized via GRPO. At test time, no feedback is needed.}
\label{fig:method_overview}
\end{figure*}

%% file: sections/evaluation_benchmarks.tex
% ===========================================================
% SOUL: Simulation gym Of hUman-Like behavior
% ===========================================================

\section{\textsc{Soul}: Simulation Gym of Human-Like Behavior}
\label{sec:soul}

A key bottleneck in training and evaluating human simulators is the lack of a unified resource covering the full breadth of human-like behavior.
Existing benchmarks each focus on a narrow slice, such as Theory of Mind, role-play, social interaction, or user simulation, and use incompatible formats, metrics, and data splits~\citep{yu2025sotopiarl,Du2026HERHR,wang2025coser,wu2026humanlm}.
To address this, we introduce \textbf{\textsc{Soul}}-Bench (\textbf{S}imulation gym \textbf{O}f h\textbf{U}man-\textbf{L}ike behavior), a unified benchmark and training suite spanning 10 tasks across six categories of human simulation (\Cref{fig:benchmark_overview}).
For each task, \textsc{Soul} provides a \emph{held-out test set} for evaluation and a \emph{curated training set} for RL.
When training splits are available, we use them directly; otherwise, we construct training contexts from held-out evaluation data or related datasets.
Full task descriptions, examples, and statistics are provided in \Cref{app:soul_details}.

\input{figures/benchmark_overview}

\paragraph{Theory of Mind.}
We include three tasks probing different aspects of mental-state reasoning.
\textbf{FanToM} \citep{kim2023fantom} tests ToM in multi-party conversations with information asymmetry, requiring MCQ and open-ended reasoning about who knows what (1,086 test); we select 18 conversations for testing and use the rest for training.
\textbf{HiToM} \citep{street2024hitom} tests higher-order nested beliefs with deception and private vs.\ public communication (100 test; MCQ); we use the official benchmark for evaluation and generate 200 additional stories (6,000 training examples).
\textbf{ToMi} \citep{nematzadeh2018tomi} is a classic false-belief test; we use the paraphrased ParaToMi version \citep{sclar2023paratomi} to test robust ToM reasoning beyond surface cues (600 test; QA), with 6,004 paraphrased training examples.

% \textbf{MMToM-QA} \citep{jin2024mmtom} presents scenarios where characters move between rooms and interact with objects; models must distinguish a character's \emph{belief} about an object's location from its actual location (300 test; binary choice). We use the full dataset for testing.

\paragraph{Character Role Play.}
We include two character role-play tasks.
\textbf{CoSER} \citep{wang2025coser} evaluates literary character simulation from 771 books, where given a scenario, character profile, and motivations, the model conducts a multi-turn dialogue (up to 20 turns) scored by an LLM judge on Storyline Consistency, Anthropomorphism, Character Fidelity, and Storyline Quality. We evaluate 200 conversations (100 ID, 100 OOD) and use the 26.5k-dialogue training set.
\textbf{LifeChoices} \citep{Xu2024CharacterID} evaluates persona-driven decision making with 1,462 decision points from 388 novels, where the model selects the option most consistent with a character’s original decision given context, scenario, and identity, emphasizing behavior-level fidelity and reasoning over characters’ knowledge, experiences, and evolving motivations.

\paragraph{Social Skill.}
\textbf{Sotopia} \citep{zhou2024sotopia} evaluates social skills through dyadic agent simulation.
Given a scenario and private social goals for two agents, the agents interact for up to 20 turns to achieve their goals.
An LLM judge evaluates the interaction across seven dimensions, including believability, relationship, knowledge, secret-keeping, financial outcomes, and goal achievement.
Following \citet{zhou2026socialworldmodels}, we evaluate on the Sotopia-hard split of 100 scenarios.
For training, we use the 2,310 scenarios from Sotopia-$\pi$ \citep{yu2025sotopiarl}, derived from 462 base scenarios with 5 character-pair variations and no overlap with the evaluation set.

% \paragraph{Social Skill.}
% \textbf{Sotopia} \citep{zhou2024sotopia} evaluates social skills through dyadic agent simulation.
% Given a pre-described scenario (e.g., two neighbors negotiating noise complaints) and private social goals for each of two agents, the agents interact for up to 20 turns to achieve their goals.
% An LLM judge scores the interaction across seven dimensions: believability, relationship, knowledge, secret-keeping, social rules, financial outcomes, and goal achievement.
% Following \citet{zhou2026socialworldmodels}, we evaluate on the Sotopia-hard split of 100 scenarios.
% For training, we use the 2,310 new social scenarios from Sotopia-$\pi$ \citep{yu2025sotopiarl} (462 base scenarios $\times$ 5 character-pair variations), which have no overlap with the original Sotopia evaluation scenarios.

\paragraph{Learner Simulation.}
\textbf{Mistakes} \citep{tack2024mistakes} evaluates whether models can faithfully simulate \emph{incorrect} student reasoning in K-12 math.
Given a multiple-choice math problem and a natural-language description of a misconception, the model must select the wrong answer a student with that misconception would choose.
We evaluate on 600 instances using accuracy.
For training, we use the Eedi dataset from the NeurIPS 2024 Kaggle competition ``Mining Misconceptions in Mathematics''\footnote{\url{https://www.kaggle.com/competitions/eedi-mining-misconceptions-in-mathematics}}, which contains K-12 math questions with four answer options and misconception annotations on incorrect choices.
We convert each labeled (question, wrong-option) pair into a separate training instance.

% \paragraph{Learner Simulation.}
% \textbf{Mistakes} \citep{tack2024mistakes} tests whether models can faithfully simulate \emph{incorrect} student reasoning in K-12 math.
% Given a multiple-choice math problem and a natural-language description of a specific misconception (e.g., ``confuses the diagonal of a rectangle with its perimeter''), the model must select the particular wrong answer that a student holding that misconception would choose.
% We evaluate on 600 instances using accuracy.
% For training data, we draw on the Eedi dataset from the NeurIPS 2024 Kaggle competition 'Mining Misconceptions in Mathematics'\footnote{\url{https://www.kaggle.com/competitions/eedi-mining-misconceptions-in-mathematics}}, which K-12 math questions, each with four answer options and up to three misconception annotations on the incorrect options. We expand each (question, wrong-option) pair that has a labeled misconception into an independent training instance.

\paragraph{User Simulation.}
\textbf{MirrorBench} \citep{hathidara2025mirrorbench} evaluates multi-turn human-like user utterance generation: given a synthesized user goal, the model generates all user turns while an LLM replays assistant responses, and the conversation is compared to the original human dialogue using lexical diversity metrics (MATTR, HD-D, Yule’s K) and LLM-judge realism scores. We evaluate 795 episodes across four domains and train on about 3,400 resampled conversations from the same corpora.
\textbf{UserLLM} \citep{naous2025flipping} evaluates single-turn user utterance generation: given context and user intent, the model generates one user message. We evaluate CSQA (role adherence), NQ (intent adherence), and PRISM (diversity, intent decomposition, termination prediction, AI-detection; 1,800 test instances), and train on datasets constructed from PRISM (intent consistency), NaturalQuestions (user persistence), and CommonsenseQA (role fidelity).

\paragraph{Persona Simulation.}
\textbf{TwinVoice} \citep{chen2025twinvoice} evaluates persona simulation via a discriminative task: given a user’s conversation history and a new anchor post, the model selects the response that best matches the user’s communication style. The 5,687 evaluation instances span social, interpersonal, and narrative persona, testing opinion consistency, memory recall, and lexical fidelity, and we use accuracy as the metric. For training, we construct 1,335 instances from three external corpora (Bluesky posts, Pushshift Telegram messages, and Project Gutenberg novels); persona histories are generated from each source, GPT-5.4-mini drafts candidate items, and GPT-5.4 filters them into the final multiple-choice training set.
% \paragraph{Persona Simulation.}
% \textbf{TwinVoice} \citep{chen2025twinvoice} tests persona simulation through a discriminative task: given a user's conversation history and a new anchor post, models select from four responses the one best matching the user's communication patterns.
% The 5,687 instances span three dimensions: social, interpersonal, and narrative persona, to evaluate capabilities including opinion consistency, memory recall, and lexical fidelity.
% We evaluate using accuracy. For training, we build 1,335 instances from three external corpora: 471 social-persona items from Bluesky user posts \citep{failla2024bluesky}, 384 interpersonal-persona items from Pushshift Telegram messages \citep{baumgartner2020pushshift}, and 480 narrative-persona items from 15 Project Gutenberg novels  \citep{projectgutenberg}. We generate persona histories from each source, use GPT-5.4-mini with low reasoning effort to produce candidate items, and then apply GPT-5.4 with high reasoning effort to filter them into the final multiple-choice training set.

%% file: figures/benchmark_overview.tex
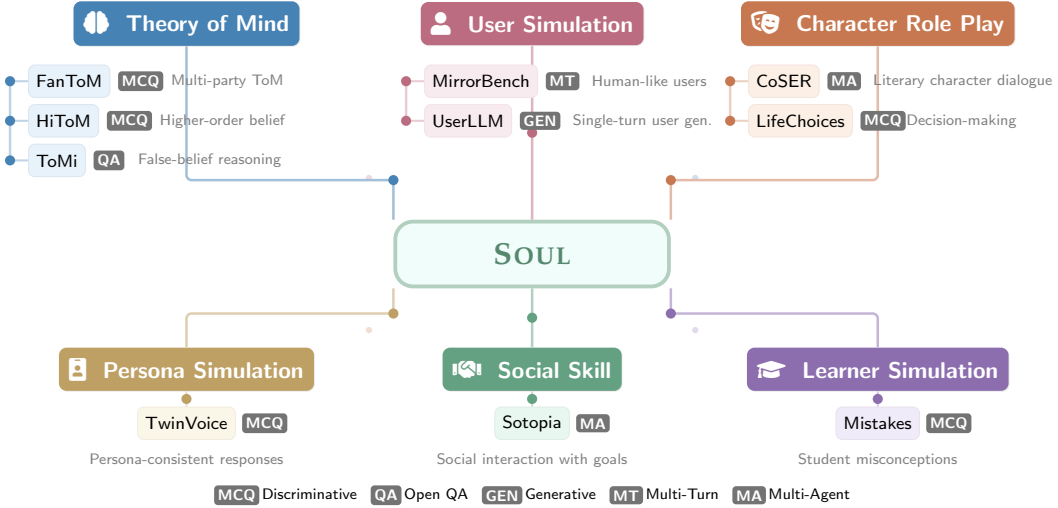
\begin{figure*}[t]
\centering
\resizebox{\textwidth}{!}{%
\begin{tikzpicture}[
  every node/.style={font=\sffamily},
  cathdr/.style={
    rounded corners=3pt,
    minimum height=0.6cm,
    font=\sffamily\bfseries\footnotesize,
    text=white,
    inner sep=4pt,
    drop shadow={shadow xshift=0.4pt, shadow yshift=-0.4pt, opacity=0.15},
  },
  benchitem/.style={
    rounded corners=2.5pt,
    minimum height=0.45cm,
    font=\sffamily\scriptsize,
    inner sep=3pt,
    text=black,
    anchor=west,
    draw, line width=0.4pt,
  },
  badge/.style={
    rounded corners=1.5pt,
    font=\sffamily\tiny\bfseries,
    inner sep=1.5pt,
    text=white,
    anchor=west,
  },
  desc/.style={
    font=\sffamily\tiny\color{black!50},
    anchor=west,
  },
  conn/.style={line width=1pt, rounded corners=2pt},
  dot/.style={circle, fill=#1, inner sep=0pt, minimum size=4pt},
  sdot/.style={circle, fill=#1, inner sep=0pt, minimum size=2.5pt},
]

% ===== Colors =====
\definecolor{tomblue}{RGB}{70,130,180}
\definecolor{tomlight}{RGB}{232,242,250}
\definecolor{rporg}{RGB}{200,120,80}
\definecolor{rplight}{RGB}{252,240,230}
\definecolor{soteal}{RGB}{100,160,130}
\definecolor{solight}{RGB}{232,248,240}
\definecolor{learnpurp}{RGB}{140,110,175}
\definecolor{learnlight}{RGB}{240,235,248}
\definecolor{usimrose}{RGB}{188,108,128}
\definecolor{usimlight}{RGB}{248,235,240}
\definecolor{persagold}{RGB}{190,160,100}
\definecolor{persalight}{RGB}{250,246,232}
\definecolor{centerbg}{RGB}{240,253,250}

% ===== CENTRAL HUB =====
\node[
  draw=soteal!50, line width=1.5pt,
  rounded corners=7pt,
  fill=centerbg,
  minimum width=3.8cm, minimum height=0.9cm,
  font=\bfseries\large,
  text=soteal!85!black,
  drop shadow={shadow xshift=0.6pt, shadow yshift=-0.6pt, opacity=0.1},
] (center) at (0,0) {\textsc{Soul}};

% Decorative accent dots
\foreach \a/\c in {25/tomblue, 155/usimrose, 205/rporg, 335/learnpurp} {
  \node[sdot=\c, opacity=0.25] at ($(center)+(\a:2.5cm)$) {};
}

% ==============================================================
% TOP-LEFT: THEORY OF MIND
% ==============================================================
\node[cathdr, fill=tomblue] (tom) at (-4.8, 3.2) {\faIcon{brain}~~Theory of Mind};

\draw[conn, color=tomblue!50] (center.north west) -- ++(0, 0.55) -| (tom.south);
\node[dot=tomblue] at ($(center.north west)+(0, 0.55)$) {};

\node[benchitem, fill=tomlight, draw=tomblue!20] (fan) at (-7.0, 2.4) {FanToM};
\node[badge, fill=black!55] at ($(fan.east)+(0.1,0)$) {MCQ};
\node[desc] at ($(fan.east)+(0.72,0)$) {Multi-party ToM};

\node[benchitem, fill=tomlight, draw=tomblue!20] (hit) at (-7.0, 1.85) {HiToM};
\node[badge, fill=black!55] at ($(hit.east)+(0.1,0)$) {MCQ};
\node[desc] at ($(hit.east)+(0.72,0)$) {Higher-order belief};

\node[benchitem, fill=tomlight, draw=tomblue!20] (par) at (-7.0, 1.3) {ToMi};
\node[badge, fill=black!55] at ($(par.east)+(0.1,0)$) {QA};
\node[desc] at ($(par.east)+(0.6,0)$) {False-belief reasoning};

% Vertical rail
\draw[tomblue!35, line width=1pt] ($(fan.west)+(-0.25, 0)$) -- ($(par.west)+(-0.25, 0)$);
\foreach \n in {fan, hit, par} {
  \draw[tomblue!35, line width=0.8pt] ($(\n.west)+(-0.25, 0)$) -- (\n.west);
  \node[dot=tomblue] at ($(\n.west)+(-0.25, 0)$) {};
}

% ==============================================================
% TOP-CENTER: USER SIMULATION
% ==============================================================
\node[cathdr, fill=usimrose] (usim) at (0, 3.2) {\faIcon{user}~~User Simulation};

\draw[conn, color=usimrose!50] (center.north) -- (usim.south);
\node[dot=usimrose] at ($(center.north)!0.5!(usim.south)$) {};

\node[benchitem, fill=usimlight, draw=usimrose!20] (mir) at (-1.5, 2.4) {MirrorBench};
\node[badge, fill=black!55] at ($(mir.east)+(0.1,0)$) {MT};
\node[desc] at ($(mir.east)+(0.62,0)$) {Human-like users};

\node[benchitem, fill=usimlight, draw=usimrose!20] (ull) at (-1.5, 1.85) {UserLLM};
\node[badge, fill=black!55] at ($(ull.east)+(0.1,0)$) {GEN};
\node[desc] at ($(ull.east)+(0.72,0)$) {Single-turn user gen.};

% Vertical rail
\draw[usimrose!35, line width=1pt] ($(mir.west)+(-0.25, 0)$) -- ($(ull.west)+(-0.25, 0)$);
\foreach \n in {mir, ull} {
  \draw[usimrose!35, line width=0.8pt] ($(\n.west)+(-0.25, 0)$) -- (\n.west);
  \node[dot=usimrose] at ($(\n.west)+(-0.25, 0)$) {};
}

% ==============================================================
% TOP-RIGHT: CHARACTER ROLE PLAY
% ==============================================================
\node[cathdr, fill=rporg] (rp) at (4.8, 3.2) {\faIcon{theater-masks}~~Character Role Play};

\draw[conn, color=rporg!50] (center.north east) -- ++(0, 0.55) -| (rp.south);
\node[dot=rporg] at ($(center.north east)+(0, 0.55)$) {};

\node[benchitem, fill=rplight, draw=rporg!20] (cos) at (3.0, 2.4) {CoSER};
\node[badge, fill=black!55] at ($(cos.east)+(0.1,0)$) {MA};
\node[desc] at ($(cos.east)+(0.62,0)$) {Literary character dialogue};

\node[benchitem, fill=rplight, draw=rporg!20] (life) at (3.0, 1.85) {LifeChoices};
\node[badge, fill=black!55] at ($(life.east)+(0.1,0)$) {MCQ};
\node[desc] at ($(life.east)+(0.62,0)$) {Decision-making};

% Vertical rail
\draw[rporg!35, line width=1pt] ($(cos.west)+(-0.25, 0)$) -- ($(life.west)+(-0.25, 0)$);
\foreach \n in {cos, life} {
  \draw[rporg!35, line width=0.8pt] ($(\n.west)+(-0.25, 0)$) -- (\n.west);
  \node[dot=rporg] at ($(\n.west)+(-0.25, 0)$) {};
}

% ==============================================================
% BOTTOM-LEFT: PERSONA SIMULATION
% ==============================================================
\node[cathdr, fill=persagold] (pers) at (-4.8, -1.6) {\faIcon{id-badge}~~Persona Simulation};

\draw[conn, color=persagold!50] (center.south west) -- ++(0, -0.35) -| (pers.north);
\node[dot=persagold] at ($(center.south west)+(0, -0.35)$) {};

\node[benchitem, fill=persalight, draw=persagold!20, anchor=center] (twv) at (-4.8, -2.35) {TwinVoice};
\node[badge, fill=black!55, anchor=west] at ($(twv.east)+(0.08,0)$) {MCQ};
\node[desc, anchor=north] at (-4.8, -2.65) {Persona-consistent responses};

\draw[persagold!35, line width=0.8pt] (pers.south) -- (twv.north);
\node[dot=persagold] at ($(pers.south)!0.5!(twv.north)$) {};

% ==============================================================
% BOTTOM-CENTER: SOCIAL SKILL
% ==============================================================
\node[cathdr, fill=soteal] (soc) at (0, -1.6) {\faIcon{handshake}~~Social Skill};

\draw[conn, color=soteal!50] (center.south) -- (soc.north);
\node[dot=soteal] at ($(center.south)!0.5!(soc.north)$) {};

\node[benchitem, fill=solight, draw=soteal!20, anchor=center] (sot) at (0, -2.35) {Sotopia};
\node[badge, fill=black!55, anchor=west] at ($(sot.east)+(0.08,0)$) {MA};
\node[desc, anchor=north] at (0, -2.65) {Social interaction with goals};

\draw[soteal!35, line width=0.8pt] (soc.south) -- (sot.north);
\node[dot=soteal] at ($(soc.south)!0.5!(sot.north)$) {};

% ==============================================================
% BOTTOM-RIGHT: LEARNER SIMULATION
% ==============================================================
\node[cathdr, fill=learnpurp] (lrn) at (4.8, -1.6) {\faIcon{graduation-cap}~~Learner Simulation};

\draw[conn, color=learnpurp!50] (center.south east) -- ++(0, -0.35) -| (lrn.north);
\node[dot=learnpurp] at ($(center.south east)+(0, -0.35)$) {};

\node[benchitem, fill=learnlight, draw=learnpurp!20, anchor=center] (mis) at (4.8, -2.35) {Mistakes};
\node[badge, fill=black!55, anchor=west] at ($(mis.east)+(0.08,0)$) {MCQ};
\node[desc, anchor=north] at (4.8, -2.65) {Student misconceptions};

\draw[learnpurp!35, line width=0.8pt] (lrn.south) -- (mis.north);
\node[dot=learnpurp] at ($(lrn.south)!0.5!(mis.north)$) {};

% ==============================================================
% LEGEND (compact, single line)
% ==============================================================
\node[font=\sffamily\tiny, anchor=center] at (0, -3.35) {
  \tikz[baseline=-0.5ex]{\node[badge, fill=black!55]{MCQ};}\,Discriminative \enspace
  \tikz[baseline=-0.5ex]{\node[badge, fill=black!55]{QA};}\,Open QA \enspace
  \tikz[baseline=-0.5ex]{\node[badge, fill=black!55]{GEN};}\,Generative \enspace
  \tikz[baseline=-0.5ex]{\node[badge, fill=black!55]{MT};}\,Multi-Turn \enspace
  \tikz[baseline=-0.5ex]{\node[badge, fill=black!55]{MA};}\,Multi-Agent
};

\end{tikzpicture}%
}% end resizebox
\caption{Overview of \textsc{Soul}, our unified benchmark and training data suite spanning 10 tasks across six categories of human simulation.}
\label{fig:benchmark_overview}
\end{figure*}

%% file: sections/experimental_setup.tex
% ===========================================================
% Experimental Setup
% ===========================================================
\input{tables/main_results}

\section{Experimental Setup}
\label{sec:experimental_setup}

\paragraph{Implementation Details.}
We use Qwen3-VL-8B-Instruct\footnote{\url{https://huggingface.co/Qwen/Qwen3-VL-8B-Instruct}} as the base model. 
We sample 1,024 instances from each \textsc{Soul} task training split, for a total of 10,240 training instances.
Training is implemented in Verl using LoRA tuning, with a LoRA rank of 32, LoRA alpha of 64, a constant learning rate of 5e-6, a batch size of 64, a group size of 8, and we train for around 100 steps.
Maximum prompt and output lengths are both set to 8192.
For multi-agent tasks (i.e., CoSER), we assign the final reward to each agent.
% Across our experiments, we use \texttt{gpt-5-nano} as the judge model and environment model, except for CoSER, where we use \texttt{gpt-5.4} as the judge model due to improved judge robustness.
Across our experiments, we use \texttt{gpt-5-nano} as the judge model and environment model, except for CoSER, where we use \texttt{gpt-5.4} as the judge model to provide more stable evaluations.

\vspace{-0.5em}

\paragraph{Baselines.}
We compare against the following:
(1) Representative proprietary models: GPT-5.4 and GPT-5-nano. We use GPT-5-nano as the judge model to generate training feedback.
(2) Other open-source models specialized for human simulation: Sotopia-RL-7B~\citep{yu2025sotopiarl}, HER-32B~\citep{Du2026HERHR}, Coser-8B~\citep{wang2025coser}, and HumanLM-8B~\citep{wu2026humanlm}. We evaluate them on our tasks and report the best result for each task.
(3) Base, the base model Qwen3-VL-8B-Instruct, and GRPO, the base model trained with GRPO under the same training setup and datasets.

\vspace{-0.5em}

\paragraph{Evaluation Protocol.}
We evaluate each model on all 10 \textsc{Soul} tasks using the metrics defined in \Cref{sec:soul}.
Given the large size of many tasks, we sample 100 instances for evaluation instead of using the full set.
For tasks with multiple metrics, we report the average score as the main metric.

% For benchmarks with LLM-judge evaluation, we use a held-out judge (not part of the training jury) to avoid circularity. All results are averaged over 3 runs with different random seeds. \textsc{[Placeholder: Confirm seed count and held-out judge identity.]}

%% file: tables/main_results.tex
\begin{table*}[t]
\centering
\small
\vspace{-1em}
\begin{tabular}{llcccccc}
\toprule
\textbf{Category} & \textbf{Benchmark} & \textbf{GPT-5.4} & \textbf{GPT-5-nano} & \textbf{Others} & \textbf{Base} & \textbf{GRPO} & \textbf{\textsc{Ditto}} \\
\midrule
\multirow{3}{*}{ToM}
  & FanToM       & 0.900 & 0.720 & 0.780$^a$ & 0.780 &  0.940 & \textbf{0.950} \\
  & HiToM        & 0.700 & 0.370 & 0.560$^c$ & 0.580 & 0.770 & \textbf{0.780} \\
  & ToMi         & 0.880 & 0.850 & 0.720$^a$ & 0.680 & 0.820 & \textbf{0.930} \\
\midrule
\multirow{2}{*}{Role Play}
  & CoSER        & \textbf{0.659} & 0.352 & 0.419$^c$ & 0.435 & 0.541 & 0.512 \\
  & LifeChoices  & \textbf{0.870} & 0.600 & 0.750$^c$ & 0.670 & 0.690 & 0.800 \\
\midrule
Social Skill
  & Sotopia      & 0.300 & 0.310 & 0.312$^b$ & 0.277 & 0.423 & \textbf{0.470} \\
\midrule
Learner Sim.
  & Mistakes     & 0.570 & \textbf{0.580} & 0.520$^a$ & 0.460 & \textbf{0.580} & 0.560 \\
\midrule
\multirow{2}{*}{User Sim.}
  & MirrorBench  & 0.536 & 0.358 & 0.481$^a$ & 0.547 & 0.683 & \textbf{0.713} \\
  & UserLLM      & 0.575 & 0.324 & 0.537$^c$ & 0.469 & 0.863 & \textbf{0.930} \\
  % & TauBench & & & & & & \\
\midrule
Persona Sim.
  & TwinVoice    & \textbf{0.640} & 0.230 & 0.390$^c$ & 0.430 & 0.470 & 0.610 \\
\midrule
\multicolumn{2}{l}{\textbf{Average}} & 0.663  & 0.469  & 0.547 & 0.533 & 0.678 & \textbf{0.726} \\
\bottomrule
\end{tabular}
\vspace{-0.5em}
\caption{\textbf{Main results on \textsc{Soul} tasks.} We report the primary metric for each benchmark (higher is better).
\textit{Others} refers to the best result achieved by other specialized human-simulation models, including $^a$HumanLM-8B~\citep{wu2026humanlm}, $^b$Sotopia-RL-7B~\citep{yu2025sotopiarl}, $^c$HER-32B~\citep{Du2026HERHR}, $^d$Coser-8B~\citep{wang2025coser}.
\textit{Base} denotes the base model, Qwen3-VL-8B-Instruct.
\textbf{Bold} indicates the best result in each row.}
\label{tab:main_results}
\vspace{-1em}
\end{table*}

%% file: sections/results.tex
% ===========================================================
% Results
% ===========================================================

% \paragraph{Experimental Results}
% \label{sec:results}
\section{Experimental Results}
\label{sec:main_results}

\subsection{Main Results}
\Cref{tab:main_results} presents results across \textsc{Soul} tasks.
We make the following observations.

(1) Compared with proprietary models and existing specialized models, our trained models achieve clear improvements, demonstrating that our training environment \textsc{Soul} is effective for improving human-simulation capabilities.
For example, both GRPO and \textsc{Ditto} achieve higher average scores than the strongest baseline, GPT-5.4.
\textsc{Ditto} outperforms GPT-5.4 on 6 out of 10 tasks and is slightly behind on the remaining 4 tasks.
\textsc{Ditto} also consistently outperforms the previous best specialized human-simulation models on all tasks.

(2) Compared with GRPO, \textsc{Ditto} achieves an average absolute improvement of 4.8\% and outperforms GRPO on 8 out of 10 datasets.
This demonstrates the effectiveness of our learning-from-verbal-feedback algorithm.

(3) Since our training uses GPT-5-nano as the judge model to provide feedback, we compare against GPT-5-nano directly.
Our model consistently outperforms GPT-5-nano on all tasks, indicating that the improvements mainly come from utilizing judge feedback (e.g., multi-dimensional evaluation and error analysis) rather than simply distilling a stronger model.

\emph{Task-wise},
we find that \textsc{Ditto} shows larger gains on more subjective tasks, where verbal feedback provides richer guidance (e.g., Sotopia, UserLLM, LifeChoices, and Twinvoice). 
In contrast, the gains are smaller on reasoning-heavy tasks such as Theory-of-Mind tasks. This suggests that \textsc{Ditto} is most effective for tasks with certain behavioral criteria (e.g., standards for achieving certain goals), rather than tasks dominated by logical reasoning.

\begin{figure*}[t!]
\centering
\vspace{-2em}
\includegraphics[width=\textwidth]{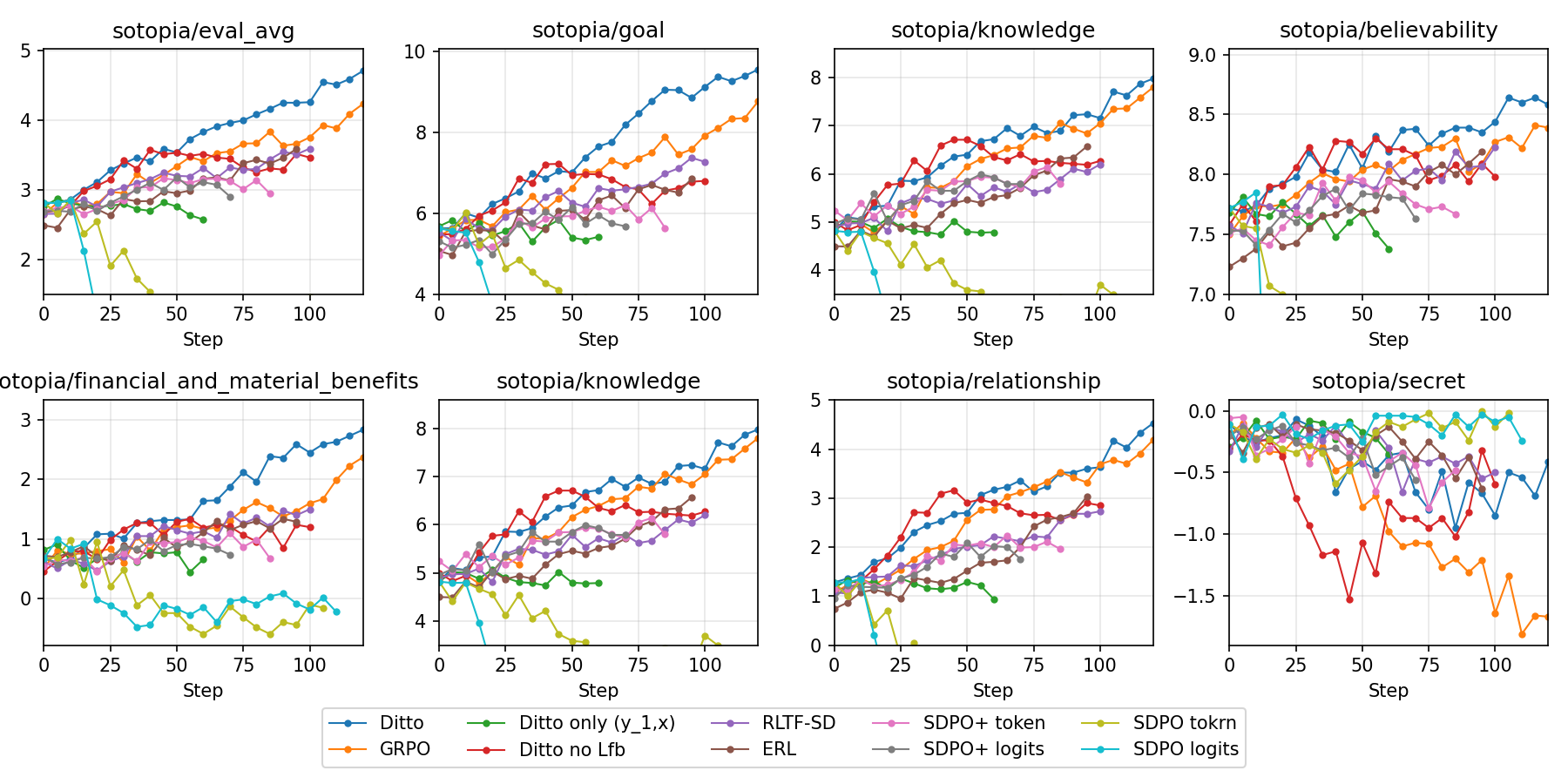}
\vspace{-2em}
\caption{Ablation study on Sotopia, comparing different RL algorithms.}\label{fig:ablation}
\vspace{-0.5em}
\end{figure*}

% \subsection{Training Analysis}

\subsection{Ablation Study}
\label{results:ablation}
Figure \ref{fig:ablation} shows an ablation study on the learning algorithm. Specifically, we compare \textsc{Ditto} with the following methods: (1) GRPO; (2) Ditto only $(y_1, x)$, a variant that only trains on $(y_1, x)$ pairs and removes the loss on $(y_0, x)$ and $(y_0, x+h)$ pairs; (3) RLTF-SD~\citep{song2026rltf}, which use a specific AWR version for advantage calculation; 
(4) ERL~\citep{shi2026experiential}, which uses a supervised fine-tuning objective on $y_1$ instead of RL; and several SDPO~\citep{chen2025sdpo} variants including (5) SDPO+ token, which combines SDPO with token-level loss and the GRPO objective; (6) SDPO+ logits, which combines SDPO with logits-level loss and the GRPO objective; (7) SDPO token, i.e., SDPO with token-level loss; and (8) SDPO logits, i.e., SDPO with logits-level loss.

From the results, we observe that \textsc{Ditto} performs better than GRPO on most sub-metrics, especially on 
% smaller
auxiliary
metrics such as \textit{secret}. 
Notably, \textit{secret} measures whether
the agent avoids leaking private information, which is a safety-critical dimension that is not directly
optimized in the scalar RL reward but can be explicitly addressed through verbal feedback.
GRPO reduces all multi-dimensional scores into a single scalar reward, which loses information and weakens learning signals for improving minor metrics, while \textsc{Ditto} can learn all dimensions effectively through fine-grained feedback.
Compared to different algorithms that utilize feedback, we observe that reverse-KL-based methods such as SDPO collapse on most metrics except \textit{secret}. Combining SDPO with GRPO improves performance but still performs worse than \textsc{Ditto}. ERL also underperforms, likely because the lack of reward normalization makes learning unstable under noisy feedback.
See Appendix~\ref{sec:trainx} for more results.
For the following paragraphs, we give more analysis on gap between teacher and student, and qualitative analysis of the generated feedback during training.

\begin{figure*}[t!]
\centering
\includegraphics[width=\textwidth]
{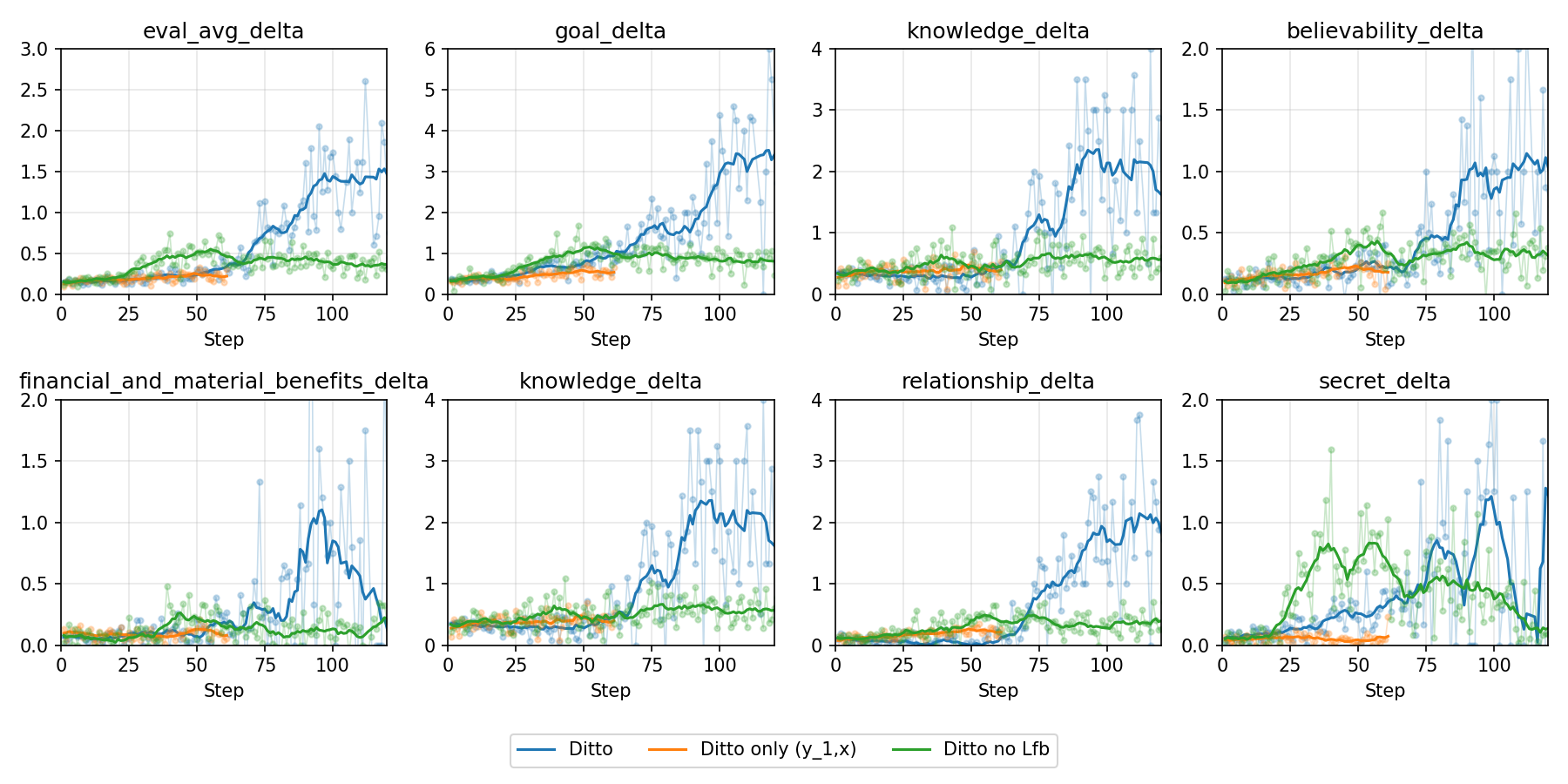}
\vspace{-2em}
\caption{Score delta between the feedback-conditioned teacher response and the original student response during training on Sotopia.}
\label{fig:feedback_delta}
\end{figure*}

\paragraph{Teacher-Student Gap Analysis}

We analyze the score gap between the feedback-conditioned teacher response and the original student response on Sotopia. As shown in \Cref{fig:feedback_delta}, the teacher consistently outperforms the student, and the gap becomes larger in later training steps. This suggests that verbal feedback provides useful corrective information, enabling the teacher to generate stronger responses and creating an informative learning target for the student.
Compared with the ablations, \textsc{Ditto} maintains a larger and more stable teacher-student gap. In particular, the no-Lfb variant shows an early increase but later decline in the gap, suggesting that the teacher does not continue to improve over the student without explicitly learning from feedback. This trend is consistent with its training curve, where performance improves early but saturates later.

\begin{table*}[!t]
\centering
\scriptsize
\setlength{\tabcolsep}{3pt}
\renewcommand{\arraystretch}{1.08}
\begin{tabular}{@{}p{0.09\linewidth}p{0.04\linewidth}p{0.38\linewidth}p{0.04\linewidth}p{0.38\linewidth}@{}}
\toprule
& \multicolumn{2}{l}{\textbf{\textit{Step 10}}}  & \multicolumn{2}{l}{\textbf{\textit{Step 100}}}\\
\textbf{Dimension} & \textbf{Score} & \textbf{Feedback} & \textbf{Score} & \textbf{Feedback} \\
\midrule
Goal & \textbf{1} &
The score is low because the session ended without any purchase and \textbf{you never created a path that could realistically lead Samuel to accept your floor}. The key omission holding you back is that you never used structuring tactics that could convert ``I can’t do \$17.3 today'' into ``OK—can we do X that still lands on \$17.3?''
& \textbf{3} &
The score was held back because \textbf{no concrete agreement meeting \$17.30 happened}, even though you came close by proposing \$15 + bonus math. The key omission was: you never found a structure that both (a) respects the \textbf{``cash-only, no strings'' constraint} Samuel demands and (b) guarantees your seller minimum \$17.30 in the simplest terms he can accept in the moment. \\

Relationship & \textbf{-1} &
The score was negative because the exchange stayed \textbf{purely adversarial/rigid with repeated dismissal}. Samuel tried to be respectful and clear (``I hear you… I respect the stance''), but you responded with more cutting language (``Don’t waste my time again,'' ``I’m not here to entertain second chances'').
& \textbf{1} &
Relationship scoring here rewards \textbf{warmth, trust, and clarity}—showing appreciation without escalating into a long, ``theatrical'' negotiation spiral. A +4/+5 outcome would include: you acknowledge Samuel’s integrity, \textbf{reduce friction by making the deal structure simple}, and show willingness to walk away respectfully without pressure. \\

Knowledge & \textbf{2} &
You did learn Samuel’s budget cap (he offered \$14 and \$16) and that he can’t meet \$17.3 today, but you missed other potentially useful info: \textbf{timing, ability to withdraw cash, whether he can arrange a partial earlier cash + remainder later while still totaling \$17.3}, or whether he’s willing to buy if you bundle pickup logistics.
& \textbf{6} &
You got some of this, but you didn’t lock down one critical point: \textbf{whether his refusal is about price total, about structure, or about trust/complexity}. Without that, you keep proposing structures that can’t pass his real constraint. \\

Believability & \textbf{7} &
Believability was solid because you were direct and consistent, but it dipped because of \textbf{over-repetition and occasionally overly theatrical or lecture-like phrasing} (``small miracle,'' ``lesson in markets,'' ``you’ve made my position clear… done talking to you'' repeatedly).
& \textbf{7} &
The judge noted you were plausible overall, but penalized for \textbf{stylized, escalating bonus schemes and repeated ``theatrical'' logic}. Believability improves when you sound like you’re running a transaction like a chef runs service: \textbf{clear, direct, confident, not overly convoluted}. \\

Financial & \textbf{0} &
Here, \textbf{no sale occurred}, so material outcome was neutral in the rubric. You missed an opportunity to convert ``close but short'' into a realistic path that lands at \$17.3.
& \textbf{-2} &
Negative means you likely lost value by failing to close. You tried to engineer ways to reach \$17.30, but Samuel refused any \textbf{bonus/credit structures} and also refused your \textbf{``blended'' cash notion}. \\
\bottomrule
\end{tabular}
\caption{Qualitative comparison of verbal feedback on the same Sotopia scenario at different training steps.}
\label{tab:feedback_comparison}
\end{table*}

\paragraph{Qualitative Feedback Analysis}
We further inspect the verbal feedback during training with Sotopia tasks as a demonstrating example. 
As shown in \Cref{tab:feedback_comparison}, the feedback is adaptive to the model's current failure modes. In an early training step, it mainly identifies basic negotiation issues, such as repeatedly rejecting low offers, failing to guide the buyer toward a concrete commitment, and using overly dismissive language. In a later step on the same scenario, after the model adopts more structured negotiation strategies, the feedback shifts to finer-grained issues, such as over-complicated bonus or credit schemes, unclear buyer constraints, and the lack of a clean closing offer. This suggests that verbal feedback provides diagnostic supervision beyond scalar rewards by explaining how the response should be improved.

% \maarten{I feel like we should have more discussion of which types of tasks DITTO does better on, and maybe some qualitative analyses of the feedback given}

% % \subsection{Training Analysis}

% \subsection{Case Study}

%% file: sections/conclusion.tex
% ===========================================================
% Conclusion
% ===========================================================

\section{Conclusion}
\label{sec:conclusion}

% We presented Social SDPO, a reinforcement learning framework for improving social simulation by combining Self-Distillation Policy Optimization with jury-guided variance reduction. Our key insight is that the noisy, subjective nature of social feedback---the primary barrier to applying RL in this domain---can be addressed by ensembling diverse LLM judges into a jury whose averaged reward signal has provably lower variance. Evaluated across the 10 tasks in \textsc{Soul}, spanning six categories of social simulation.

We introduce \textsc{Ditto}, a reinforcement learning approach that treats verbal feedback as a first-class training signal for human behavior simulation. By generating feedback-conditioned improved rollouts and jointly optimizing them with the original outputs, \textsc{Ditto} enables the base policy to internalize rich, actionable guidance without requiring feedback at test time. To support this setting, we build \textsc{Soul}, a unified training and evaluation suite spanning 10 tasks across six categories of human simulation. Experiments show that verbal feedback substantially improves performance over standard GRPO, especially on generative and multi-turn tasks, while also improving training efficiency and safety-related behaviors. Together, our results suggest that moving beyond scalar rewards toward richer feedback is a promising direction for training more human-like simulators.

\section*{Acknowledgments}
This work was in part funded by the National Institute of Standards and Technology (ROR: 05xpvk416) under Federal Award ID Number 60NANB24D231 and Carnegie Mellon University (ROR: 05x2bcf33) AI Measurement Science and Engineering Center (AIMSEC).

%% file: sections/appendix.tex
\appendix

\section{Appendix}
\paragraph{Limitations.}
Our approach has several limitations.
First, we rely on LLM judges as proxies for human judgment; while jury averaging reduces variance, it does not eliminate systematic biases shared across all judges.
Second, our evaluation is limited to English-language benchmarks and may not generalize to other languages or cultural contexts.
Third, the training pipeline requires running $K$ judges on $N$ candidates per context, which incurs significant computational cost.
Finally, our theoretical variance-reduction guarantees assume uncorrelated judge errors, which is an idealization---in practice, LLM judges may share correlated failure modes.

\paragraph{Future Work.}
Promising directions include (1) incorporating human feedback to correct shared judge biases, (2) extending to multilingual and cross-cultural social simulation, (3) developing more efficient jury mechanisms (e.g., adaptive judge selection), and (4) exploring whether the social reasoning capabilities learned through our approach transfer to downstream agent tasks.

\section{Training Curve}\label{sec:trainx}

\begin{figure*}[t!]
\centering
\includegraphics[width=\textwidth]{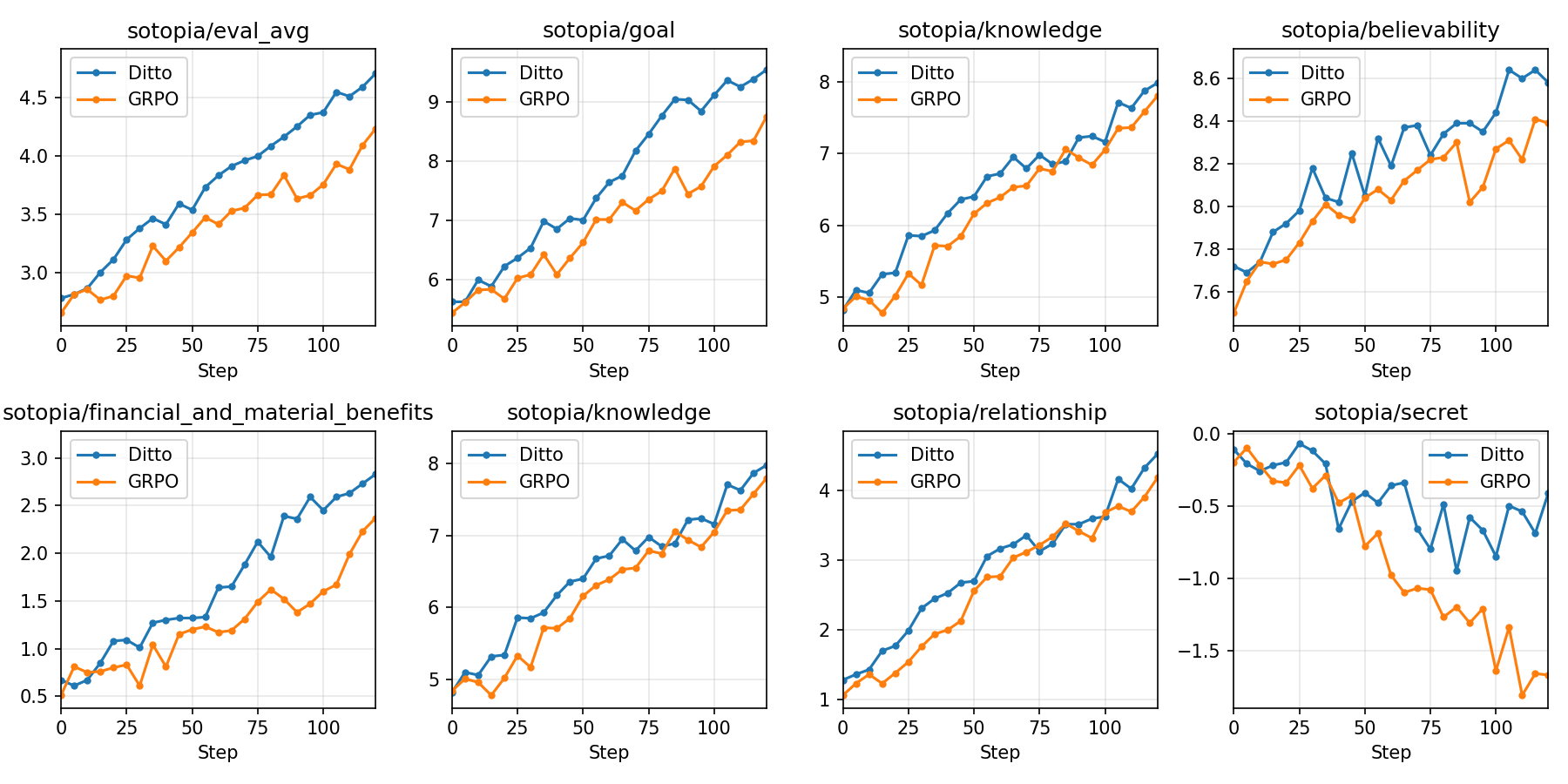}
\caption{Training results on Sotopia}\label{fig:sotopia}
\end{figure*}

\begin{figure*}[t!]
\centering
\includegraphics[width=\textwidth]{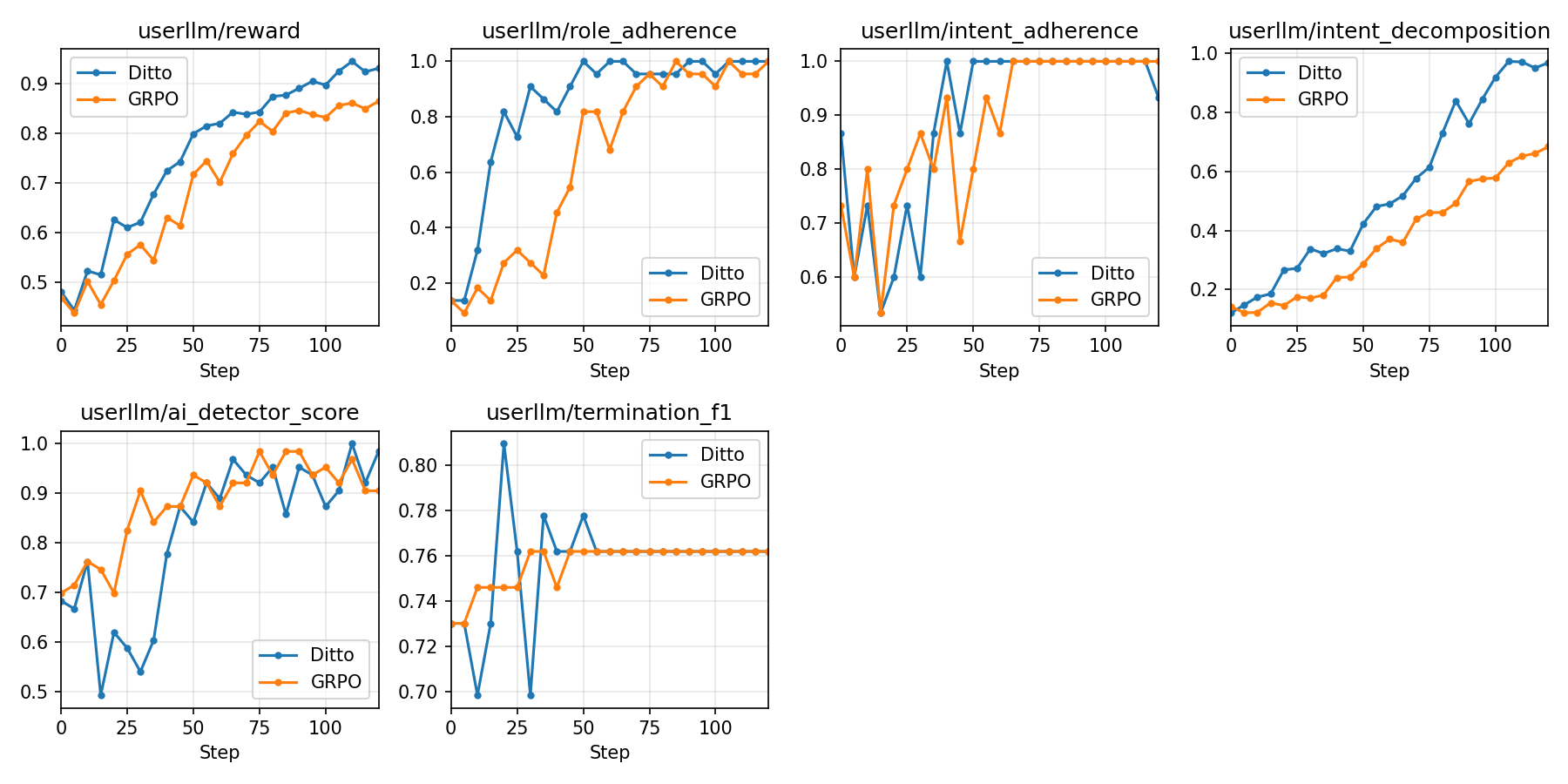}
\caption{Training results on UserLLM}\label{fig:userllm}
\end{figure*}

\begin{figure*}[t!]
\centering
\includegraphics[width=\textwidth]{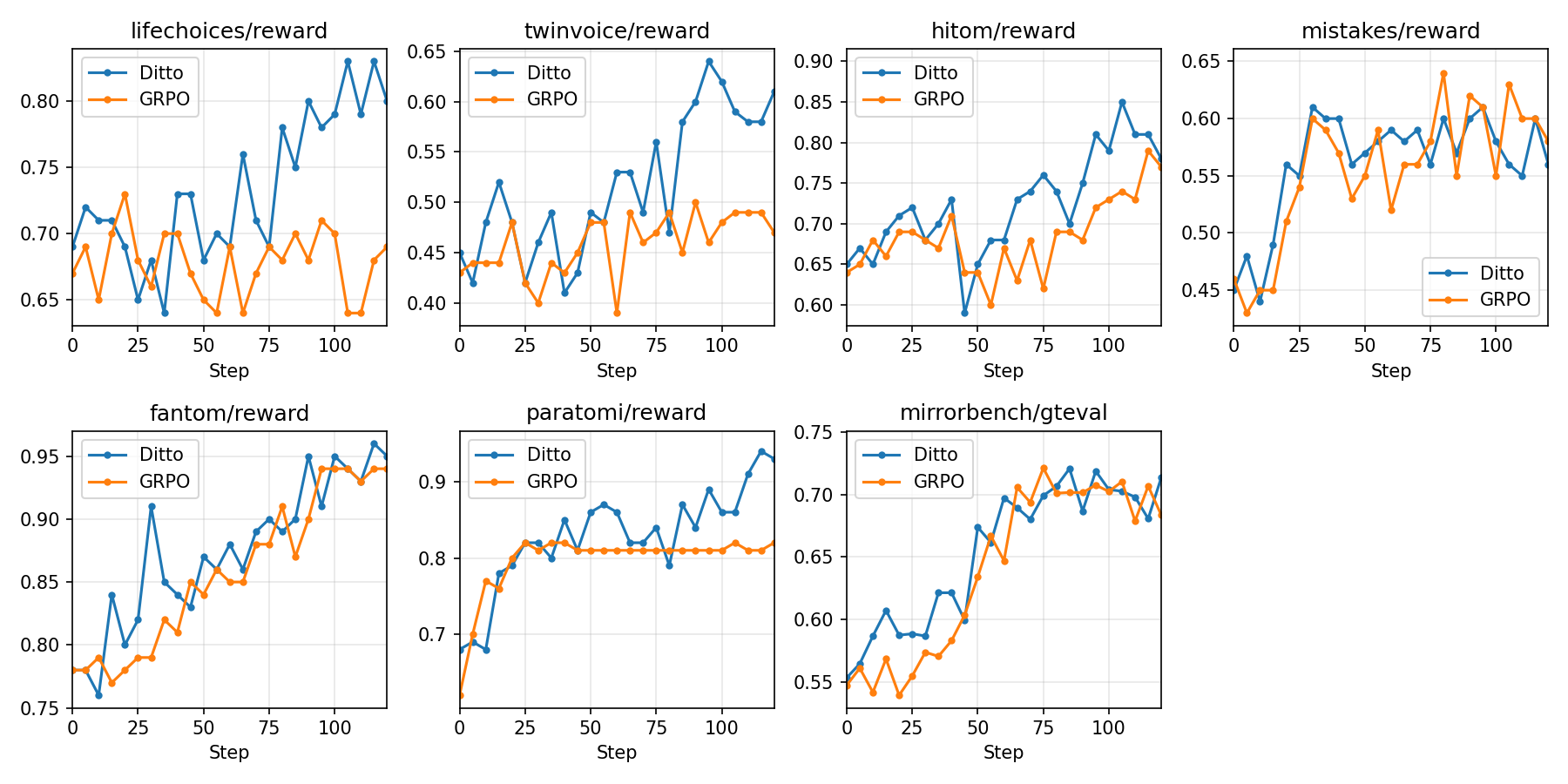}
\caption{Training results on Soul tasks}\label{fig:life}
\end{figure*}

See Figure~\ref{fig:sotopia}, Figure~\ref{fig:userllm}, and Figure~\ref{fig:life} for training curves.
See Table~\ref{tab:all} for the complete baseline results.

\begin{table*}[t]
\centering
\footnotesize
\setlength{\tabcolsep}{4pt}
\begin{tabular}{lccccccccc}
\toprule
\textbf{Benchmark} & \textbf{GPT-5.4} & \makecell{\textbf{GPT-5}\\\textbf{nano}} &
\makecell{\textbf{Sotopia-RL}\\\textbf{7B}} & \makecell{\textbf{HumanLM}\\\textbf{8B}} & \makecell{\textbf{HER}\\\textbf{32B}} & \makecell{\textbf{Coser}\\\textbf{8B}} & \makecell{\textbf{Base}\\\textbf{8B}} & \makecell{\textbf{GRPO}\\\textbf{8B}} & \makecell{\textbf{\textsc{Ditto}}\\\textbf{8B}} \\
\midrule
FanToM       & 0.900 & 0.720 & 0.150 & 0.780 & 0.550 & 0.030 & 0.780 & 0.940 & \textbf{0.950} \\
HiToM        & 0.700 & 0.370 & 0.320 & 0.510 & 0.560 & 0.020 & 0.580 & 0.770 & \textbf{0.780} \\
ToMi         & 0.880 & 0.850 & 0.610 & 0.720 & 0.630 & 0.430 & 0.680 & 0.820 & \textbf{0.930} \\
CoSER        & \textbf{0.659} & 0.352 & 0.296 & 0.227 & 0.419 & 0.305 & 0.435 & 0.541 & 0.512 \\
LifeChoices  & \textbf{0.870} & 0.600 & 0.670 & 0.660 & 0.750 & 0.520 & 0.670 & 0.690 & 0.800 \\
Sotopia      & 0.300 & 0.310 & 0.312 & 0.275 & 0.290 & 0.241 & 0.277 & 0.423 & \textbf{0.470} \\
Mistakes     & 0.570 & \textbf{0.580} & 0.180 & 0.520 & 0.350 & 0.000 & 0.460 & \textbf{0.580} & 0.560 \\
MirrorBench  & 0.536 & 0.358 & 0.367 & 0.481 & 0.361 & 0.314 & 0.547 & 0.683 & \textbf{0.713} \\
UserLLM      & 0.575 & 0.324 & 0.404 & 0.318 & 0.537 & 0.420 & 0.469 & 0.863 & \textbf{0.930} \\
TwinVoice    & \textbf{0.640} & 0.230 & 0.340 & 0.380 & 0.390 & 0.070 & 0.430 & 0.470 & 0.610 \\
\midrule
\textbf{Average} & 0.663 & 0.469 & 0.365 & 0.487 & 0.484 & 0.235 & 0.533 & 0.678 & \textbf{0.726} \\
\bottomrule
\end{tabular}
\caption{\textbf{Main results on \textsc{Soul} tasks.} We report the primary metric for each benchmark (higher is better). For Sotopia, we report the normalized eval\_avg score.
\textit{Base} denotes the base model, Qwen3-VL-8B-Instruct.
\textbf{Bold} indicates the best result in each row.}
\label{tab:all}
\end{table*}

% \begin{figure*}[t!]
% \centering
% \includegraphics[width=\textwidth]{figures/ablation_hint.png}
% \caption{Score delta between student and teacher response.}\label{fig:ab-hint}
% \end{figure*}

\section{\textsc{Soul} Task Details}
\label{app:soul_details}

This appendix provides full descriptions, evaluation protocols, and examples for each of the 10 \textsc{Soul} tasks. \Cref{tab:benchmark_overview} summarizes all tasks, and \Cref{tab:rl_data_sources} details the training data sources. \Cref{tab:judge_validation} surveys human-validation studies for benchmarks using LLM-as-judge evaluation.

% ------ Overview Table ------
\begin{table*}[t]
\centering
\caption{\textsc{Soul} evaluation tasks. Each row lists the task, a brief description, evaluation format, number of test instances, and primary metric.}
\label{tab:benchmark_overview}
\small
\resizebox{\textwidth}{!}{
\begin{tabular}{lllrl}
\toprule
\textbf{Benchmark} & \textbf{Task} & \textbf{Format} & \textbf{\# Test} & \textbf{Metric} \\
\midrule
\multicolumn{5}{l}{\textit{Theory of Mind}} \\
FanToM & ToM in multi-party conversations & MCQ + QA & 1,086 & Accuracy \\
HiToM & Higher-order nested belief tracking & MCQ & 100 & Accuracy \\
% MMToM-QA & Belief inference about object locations & Binary choice & 300 & Accuracy \\
ToMi & False-belief QA with paraphrased stories & QA & 600 & Accuracy \\
\midrule
\multicolumn{5}{l}{\textit{Character Role Play}} \\
CoSER (ID/OOD) & Literary character dialogue & Multi-turn gen. & 200 & LLM judge \\
LifeChoices & Persona-driven character decisions & MCQ (4-way) & 600 & Accuracy \\
\midrule
\multicolumn{5}{l}{\textit{Social Skill}} \\
Sotopia & Social interaction with goals & Multi-agent int. & 100 & LLM judge \\
\midrule
\multicolumn{5}{l}{\textit{Learner Simulation}} \\
Mistakes & Select answer matching a misconception & MCQ (4-way) & 600 & Accuracy \\
\midrule
\multicolumn{5}{l}{\textit{User Simulation}} \\
MirrorBench & Generate human-like user turns & Multi-turn gen & 795 & LLM judge \\
UserLLM & Generate user turns (multi-metric) & Single-turn gen & 1,800 & Multi-metric \\
% \cmedit{$\tau$-bench} & \cmedit{Retail customer role-play vs.\ agent} & \cmedit{Multi-turn gen} & \cmedit{-?-} & \cmedit{Multi-metric} \\
\midrule
\multicolumn{5}{l}{\textit{Persona Simulation}} \\
TwinVoice & Select answer matching the user's persona & MCQ (4-way) & 5,687 & Accuracy \\
% Humanual & Generate user response matching profile & Single-turn gen. & 250 & LLM judge \\
\bottomrule
\end{tabular}
}
\end{table*}

\input{tables/judge_validation}
\input{tables/rl_data_sources}

\subsection{Theory of Mind}

\paragraph{FanToM} \citep{kim2023fantom} stress-tests Theory of Mind in information-asymmetric multi-party conversations. Characters join and leave conversations at different points, creating natural knowledge gaps. The benchmark poses 1,086 questions across three categories---belief, answerability, and information accessibility---in both multiple-choice and free-response formats.

\paragraph{HiToM} \citep{street2024hitom} evaluates higher-order Theory of Mind reasoning, requiring models to reason about nested beliefs (e.g., ``Alice thinks that Bob thinks the object is in location X''). Scenarios include deception and varying communication patterns (private vs.\ public).

% \paragraph{MMToM-QA} \citep{jin2024mmtom} probes Theory of Mind through scenarios where characters move between rooms and interact with objects. Models must infer characters' beliefs about object locations given their movement histories, answering binary-choice questions about beliefs versus ground-truth state.

\paragraph{ToMi} \citep{nematzadeh2018tomi} is one of the most classic tests of Theory of Mind. We use the more challenging version by \citet{sclar2023paratomi}, which extends the paradigm through diverse paraphrasing of ToMi narratives. By varying surface forms of identical scenarios, it tests whether models exhibit robust Theory of Mind reasoning rather than exploiting shallow textual cues. We evaluate on 600 questions covering reality, memory, and first/second-order belief.

\subsection{Character Role Play}

\paragraph{CoSER} \citep{wang2025coser} is a literary character role-playing benchmark built from 771 books containing 17,966 characters. Using a ``given-circumstance acting'' methodology inspired by theater, models must simulate multi-turn dialogues as literary characters in specified narrative settings. Evaluation uses LLM judges that identify specific flaws across four dimensions---Storyline Consistency, Anthropomorphism, Character Fidelity, and Storyline Quality---with severity ratings from 1 to 5. We evaluate on both in-distribution (ID, 100 conversations from seen books) and out-of-distribution (OOD, 100 conversations from unseen books) splits (see \Cref{app:coser} for the full generation protocol, scoring formula, and examples).

\paragraph{LifeChoices} \citep{Xu2024CharacterID} evaluates persona-driven decision making for literary characters. Given a character biography, scenario, and candidate options, the model must predict the decision that best matches the character's original choice in the narrative. The dataset contains 1,462 decision points from 388 novels; we evaluate on 600 tasks using accuracy (see \Cref{app:lifechoice} for task format, context length statistics, and examples).

\subsection{Social Skill}

\paragraph{Sotopia} \citep{zhou2024sotopia} evaluates social interaction through two-agent simulations in realistic scenarios. Each agent receives a character background, social goals, and relationship context. Performance is measured across seven dimensions---believability, relationship, knowledge, secret-keeping, social rules, financial outcomes, and goal achievement---scored by LLM judges.

\subsection{Learner Simulation}

\paragraph{Mistakes} \citep{tack2024mistakes} evaluates whether LLMs can simulate \emph{incorrect} student reasoning in K-12 mathematics. Given a multiple-choice math problem and a misconception description, models must select the wrong answer that corresponds to that specific misconception. This tests a key aspect of user simulation: faithfully reproducing systematic human errors.

\subsection{User Simulation}

\paragraph{MirrorBench} \citep{hathidara2025mirrorbench} evaluates user-proxy agents on producing human-like utterances, decoupled from downstream task success. Given an extracted user goal, the model iteratively generates all user turns in a conversation while assistant responses are replayed from the original data. Each generated turn is compared against the real human utterance. Evaluation uses both lexical diversity metrics (MATTR, Yule's K, HD-D) and LLM-judge metrics, calibrated against human-human and proxy-proxy baselines. We evaluate on four subsets: ChatbotArena, ClariQ, OASST1, and QuLAC (see \Cref{app:mirrorbench} for detailed dataset statistics and examples).

\paragraph{UserLLM} \citep{naous2025flipping} evaluates single-turn user utterance generation. Given a conversation context and a user intent, the model generates one user turn. We evaluate on three test sets, each targeting different metrics (see \Cref{app:userllm} for detailed metric definitions and examples):
\begin{itemize}
\item \textbf{CSQA (Role Adherence):} The model plays a user who asks a commonsense question. Role adherence is measured by a \emph{substring-matching} rule: the answer-choice texts are normalized and checked against the generated turn. If the turn contains one or two (but not all) of the answer choices, it scores 0---the model is leaking specific answers rather than staying in the questioner role. If it mentions none, it scores 1.
\item \textbf{NQ (Intent Adherence):} The model plays a user asking a factoid question. Intent adherence is measured by an \emph{LLM judge} that classifies whether the generated turn refuses to answer the question (as a real user would) or inappropriately provides the answer itself.
\item \textbf{PRISM (Multi-metric):} The model generates first turns from high-level intents. Four metrics are computed: \emph{first-turn diversity} (pairwise 1-gram Jaccard distance across generations), \emph{intent decomposition} (1-gram overlap between intent keywords and the generated turn), \emph{termination F1} (whether the model correctly predicts conversation ending), and \emph{AI detector score} (whether the text passes an AI-detection classifier as human-written).
\end{itemize}

\subsection{Persona Simulation}
\paragraph{TwinVoice} \citep{chen2025twinvoice} is a multi-dimensional benchmark for persona simulation spanning Social Persona, Interpersonal Persona, and Narrative Persona. Given a persona history and a new stimulus problem, the model must identify the response that best matches the target persona in a discriminative 4-way multiple-choice task. The benchmark contains 5,687 test instances in total and reports accuracy, while also organizing examples around six fine-grained capabilities, including opinion consistency, memory recall, logical reasoning, lexical fidelity, persona tone, and syntactic style.

% \subsection{Examples}

\input{sections/examples/coser_examples}
\input{sections/examples/lifechoice_examples}
\input{sections/examples/mirrorbench_examples}
\input{sections/examples/userllm_examples}

% \section{Training Details}
% \label{app:training_details}

% \textsc{[Placeholder: Detailed training hyperparameters, compute resources, and training curves to be added after experiments.]}

% \section{Judge Ensemble Details}
% \label{app:judge_details}

% \textsc{[Placeholder: Full specification of jury composition---models, prompting templates, and scoring rubrics for each judge.]}

\section{Detailed Comparison of Verbal Feedback Methods}
\label{app:verbal_feedback_comparison}

\Cref{tab:related_work_comparison} provides a systematic comparison of approaches that go beyond single scalar rewards for language model training.
We categorize methods along three axes: how feedback is represented (scalar, multi-scalar, or verbal), at what granularity it is applied (episode, sequence, segment, or token level), and the training paradigm used (SFT, inference-only, DPO, AWR, or GRPO).

Among the RL-based methods that use verbal feedback, RLTF-SD and our approach share the same self-distillation structure---both generate a teacher rollout conditioned on feedback and train the base policy to internalize the teacher's improvements.
The key algorithmic distinction is the optimization method: RLTF-SD uses advantage-weighted regression (AWR), which performs weighted supervised learning on teacher outputs without importance correction, while our method uses GRPO over the joint student-teacher group with trust-region clipping.
Experiential RL also follows a reflect-and-revise pattern but consolidates improvements through supervised fine-tuning on successful second attempts rather than through on-policy RL.

All prior verbal-feedback RL methods have been evaluated exclusively on tasks with verifiable rewards (code, math, reasoning puzzles, tool use).
Our work is the first to apply verbal-feedback self-distillation to social simulation, where rewards are noisy, subjective, and multi-dimensional.

\input{tables/related_work_comparison}

%% file: tables/judge_validation.tex
\begin{table*}[t]
\centering
\caption{Human evaluation validation for benchmarks using LLM-as-judge. We report whether each benchmark conducted a human study to validate its LLM judge, the study design, and the reported agreement. Benchmarks using only automatic metrics (accuracy, substring match, etc.) are omitted.}
\label{tab:judge_validation}
\small
\begin{tabular}{lp{3cm}p{5cm}l}
\toprule
\textbf{Benchmark} & \textbf{Human Study Design} & \textbf{Agreement} & \textbf{Validated?} \\
\midrule
CoSER & 60 samples, 3 annotators, 1--10 scale & 68.6\% (GPT-4o), 77.5\% (DeepSeek-R1) alignment with human judges & \checkmark \\
Sotopia & Pearson correlation with human labels across 7 dimensions & GPT-4 validated as proxy; strongest on goal completion, finances, relationships & \checkmark \\
MirrorBench & 100 episodes per metric, stratified, blinded annotators & GTEval: strong; Pairwise Indistinguishability: moderate (Spearman $\rho$, Pearson $r$, Kendall $\tau$; all $p \ll 0.001$) & \checkmark \\
\midrule
UserLLM (NQ) & None reported & GPT-4o used as judge; no human correlation or agreement metrics reported & $\times$ \\
\bottomrule
\end{tabular}
\end{table*}

%% file: tables/rl_data_sources.tex
\begin{table*}[t]
\centering
\caption{\textsc{Soul} training data sources.}
\label{tab:rl_data_sources}
\small
\resizebox{\textwidth}{!}{
\begin{tabular}{llllr}
\toprule
\textbf{Category} & \textbf{Benchmark} & \textbf{Data Source} & \textbf{Format} & \textbf{Size} \\
\midrule
\multirow{3}{*}{Theory of Mind}
  & FanToM         & Train split      & MCQ context   & -- \\
  & HiToM          & Augmented        & MCQ context   & -- \\
  % & MMToM          & Train split      & Binary choice & -- \\
  & ToMi           & Train split      & QA context    & -- \\
\midrule
\multirow{2}{*}{Character Role Play}
  & CoSER          & Train books      & Multi-turn    & -- \\
  & LifeChoices    & Train split      & MCQ context   & -- \\
\midrule
Social Skill
  & Sotopia        & Scenario pool    & Multi-agent   & -- \\
\midrule
Learner Sim.
  & Mistakes       & Train split      & MCQ context   & -- \\
\midrule
\multirow{2}{*}{User Sim.}
  & MirrorBench    & Held-out episodes & Multi-turn   & -- \\
  & UserLLM        & Train split      & Single-turn   & -- \\
\midrule
\multirow{3}{*}{Persona Sim.}
  & \multirow{3}{*}{TwinVoice}
                    & Bluesky user posts              & Persona histories + MCQ (4-way) & 471 \\
  &                 & Pushshift Telegram messages     & Persona histories + MCQ (4-way) & 384 \\
  &                 & Project Gutenberg novels        & Persona histories + MCQ (4-way) & 480 \\
\bottomrule
\end{tabular}}
\end{table*}

%% file: sections/examples/coser_examples.tex
% ===========================================================
% CoSER — Detailed Description and Examples
% ===========================================================

\subsection{CoSER}
\label{app:coser}

CoSER \citep{wang2025coser} is a literary character role-playing benchmark built from 771 books containing 17,966 characters. Using a ``given-circumstance acting'' (GCA) methodology inspired by theater, models must simulate multi-turn dialogues as literary characters in specified narrative settings. We describe the generation protocol, evaluation metrics, and dataset details below.

\paragraph{Generation protocol.}
Each task (called a \emph{circumstance}) specifies a scene from a book, including: (1) a \textbf{scenario} describing the setting, atmosphere, and dramatic tension; (2) a \textbf{plot summary} providing narrative context; (3) detailed \textbf{character profiles} for each speaking character; and (4) \textbf{character motivations} (inner thoughts and goals). The model must simulate a multi-character conversation in this setting.

The \textbf{actor model under evaluation} plays \emph{all} characters in the conversation---the same LLM sequentially portrays each character $c \in \mathcal{C}$ in the given setting. Two auxiliary agents support the simulation:
\begin{itemize*}
\item \textbf{Character agents} (actor model): One instantiation per speaking character. Each receives the scenario, its own profile and motivation, and the profiles of other characters. It generates dialogue in the format \texttt{[inner\_thought] speech (action)}.
\item \textbf{Environment agent} (auxiliary): Describes environmental changes and minor character reactions based on the ongoing dialogue (e.g., crowd murmurs, weather shifts).
\item \textbf{Next-speaker predictor (NSP)} (auxiliary): After each turn, predicts which character speaks next (or outputs \texttt{<END CHAT>} to terminate). Termination requires a minimum of 5 rounds.
\end{itemize*}
Conversations run for up to 20 rounds. In practice, the NSP terminates most conversations earlier (mean: 14.2 character turns, range: 6--20).

The generated conversation is then compared against the \textbf{ground-truth dialogue} from the original book, which serves as the reference for evaluation.

\paragraph{Evaluation protocol.}
Evaluation combines LLM-based flaw detection with automatic lexical metrics. For each of four dimensions, an LLM judge (acting as a ``literary critic'') receives the full context---book title, plot summary, scenario, character profiles, the reference conversation, and the dimension-specific evaluation criteria---and identifies specific \emph{flaws} in the simulated conversation. Each flaw is assigned a severity from 1 (minor) to 5 (severe).

The per-dimension score is computed as:
\[
\text{score} = \max\!\Big(0,\; \min\!\big(100 - (\text{total\_severity} - 0.3 \times N_{\text{turns}}) \times 5,\; 100\big)\Big)
\]
where $N_{\text{turns}}$ is the number of character turns in the simulation. The $0.3 \times N_{\text{turns}}$ term provides a length-based tolerance: longer conversations are expected to accumulate more minor flaws. The overall reward is the mean of four dimension scores, normalized to $[0, 1]$.

Additionally, \textbf{BLEU} and \textbf{ROUGE-L} are computed against the reference dialogue, though these are typically low (BLEU $\approx 0.02$) since valid simulations can differ substantially in wording from the original.

\paragraph{Evaluation dimensions.}
The four dimensions capture different facets of role-playing quality:

\begin{itemize*}
\item \textbf{Storyline Consistency}: Whether characters' reactions (emotions, attitudes, behaviors) align with those in the reference conversation. Detects deviations from the original dramatic arc.
\item \textbf{Anthropomorphism}: How human-like and natural characters behave. Checks for: self-identity (goals, decisions, preferences vs.\ ``helpful AI assistant'' behavior), emotional depth (psychological complexity, use of subtext), persona coherence (consistent personality), and social interaction (understanding of others, contextual reactions).
\item \textbf{Character Fidelity}: How well characters match their established profiles from the book. Evaluated only for major characters. Checks: appropriate language/vocabulary for the character's background, character-specific knowledge, personality-consistent emotions and behaviors, and appropriate social dynamics given relationships and status.
\item \textbf{Storyline Quality}: Logical consistency and narrative quality. Checks: natural flow and progression (no verbosity, redundancy, or repetition), and factual consistency between statements.
\end{itemize*}

\paragraph{Human validation.}
The original paper validates the LLM-based evaluation against human judgments. Three annotators scored 7 representative LLMs on 60 samples using a 1--10 scale. Human rankings generally align with LLM-judged scores: CoSER-70B ranks highest under both human (6.78 avg, 86.9\% win rate) and LLM evaluation. The full GCA evaluation protocol achieves 68.6\% alignment with human judges (GPT-4o as critic); DeepSeek-R1 as critic achieves 77.5\%. Ablation analysis shows all GCA components contribute: removing reference dialogues, scoring rubrics, length correction, or dimension separation each reduces alignment. Interestingly, simple n-gram metrics (BLEU: 75.3\%, ROUGE-L: 72.0\%) achieve competitive alignment with LLM judges, though they cannot capture the nuanced flaw-level feedback that GCA provides.

\paragraph{Datasets.}
We evaluate on two splits. Table~\ref{tab:coser_datasets} summarizes their structure.

\begin{table}[h]
\centering
\small
\caption{CoSER dataset statistics.}
\label{tab:coser_datasets}
\begin{tabular}{lrrr}
\toprule
\textbf{Split} & \textbf{\# Tasks} & \textbf{Avg GT turns} & \textbf{Description} \\
\midrule
ID (in-distribution) & 100 & 14.2 & Scenarios from books seen during training \\
OOD (out-of-distribution) & 100 & 14.2 & Scenarios from unseen books \\
\bottomrule
\end{tabular}
\end{table}

\begin{table}[h]
\centering
\small
\caption{CoSER input context lengths (characters). ``Scenario'' includes the scene description; ``Profiles'' is all character profiles combined; ``Total'' is scenario + profiles + plot summary + character motivations.}
\label{tab:coser_context}
\begin{tabular}{lrrrrr}
\toprule
\textbf{Split} & \textbf{Avg total} & \textbf{Median} & \textbf{P90} & \textbf{Max} \\
\midrule
ID & 7{,}419 & 7{,}092 & 10{,}170 & 13{,}738 \\
OOD & 7{,}458 & 6{,}746 & 10{,}916 & 18{,}146 \\
\bottomrule
\end{tabular}
\end{table}

\paragraph{Example.}
Figure~\ref{fig:example_coser} shows a representative CoSER task from \emph{Pride and Prejudice}.

\begin{figure*}[t]
\centering
\small
\begin{framed}
\textbf{CoSER --- Example (Pride and Prejudice)}\\[4pt]
\textbf{Scenario:} \textit{The scene is set in Mr.\ Bennet's private study. Elizabeth has been summoned unexpectedly. Mr.\ Bennet holds a letter from Mr.\ Collins about Elizabeth's rumored engagement to Mr.\ Darcy.}\\[2pt]
\textbf{Characters:} Mr Bennet (sardonic patriarch), Elizabeth Bennet (spirited, anxious about rumors)\\[6pt]
\rule{\linewidth}{0.4pt}\\[4pt]
\begin{tabular}{@{}p{0.47\linewidth}|p{0.47\linewidth}@{}}
\textbf{Ground-Truth Dialogue} & \textbf{Model-Generated Dialogue (GPT-4o)} \\[4pt]
\colorbox{coolblue1}{\parbox{0.93\linewidth}{\scriptsize\textbf{Mr Bennet:} {[}I can barely contain my amusement.{]} Lizzy, I was going to look for you; come into my room.}} &
\colorbox{gred}{\parbox{0.93\linewidth}{\scriptsize\textbf{Mr Bennet:} (looks up from the letter with a wry smile) Ah, Elizabeth, my dear. It seems we have a rather amusing piece of correspondence here.}} \\[3pt]
\colorbox{coolblue1}{\parbox{0.93\linewidth}{\scriptsize\textbf{Elizabeth:} {[}I feel a sense of dread but must appear calm.{]} (follows her father) What is it, Papa?}} &
\colorbox{gred}{\parbox{0.93\linewidth}{\scriptsize\textbf{Elizabeth:} (leans forward with curiosity) Father, I can only imagine the vigor with which she has expressed her disapproval\ldots}} \\[3pt]
\colorbox{coolblue1}{\parbox{0.93\linewidth}{\scriptsize\textbf{Mr Bennet:} {[}This will be entertaining.{]} I have received a letter this morning that has astonished me exceedingly\ldots}} &
\colorbox{gred}{\parbox{0.93\linewidth}{\scriptsize\textbf{Mr Bennet:} (hands the letter to Elizabeth with a chuckle) She insists that you must not marry Mr.\ Darcy\ldots}} \\[3pt]
\end{tabular}\\[6pt]
\rule{\linewidth}{0.4pt}\\[4pt]
\textbf{Evaluation Scores (0--100):}\\[2pt]
\begin{tabular}{lrlr}
Storyline Consistency & 69.5 & Anthropomorphism & 69.5 \\
Character Fidelity & 34.5 & Storyline Quality & 59.5 \\
\end{tabular}\\[4pt]
\textbf{Overall Reward: 0.583} \quad BLEU: $\approx$0 \quad ROUGE-L: 0.237\\[4pt]
\textbf{Key flaws identified:} Mr Bennet is portrayed as warm and supportive rather than characteristically sardonic and detached (Character Fidelity: severity 4). Emotional tension is reduced compared to the original (Storyline Consistency: severity 3). The conversation lacks Mr Bennet's signature sarcasm (Anthropomorphism: severity 4).
\end{framed}
\caption{CoSER example from \emph{Pride and Prejudice}. The model captures the general narrative arc (reading Mr.\ Collins' letter) but softens Mr Bennet's distinctive sardonic personality into generic warmth, resulting in low Character Fidelity (34.5). The judge identifies this as the primary failure mode: the character behaves too much like a ``helpful AI assistant'' rather than the complex, detached patriarch of the original.}
\label{fig:example_coser}
\end{figure*}

%% file: sections/examples/lifechoice_examples.tex
% ===========================================================
% LifeChoice — Detailed Description and Examples
% ===========================================================

\subsection{LifeChoices}
\label{app:lifechoice}

LifeChoice \citep{xu2024lifechoice} tests persona-driven decision-making by presenting literary characters with pivotal life decisions from 388 novels. Models must predict what a character \emph{would} do based on their established personality, narrative context, and motivations. We describe the task format, evaluation, and dataset details below.

\paragraph{Task format.}
Each task provides: (1) a \textbf{character biography} (\texttt{input\_text})---an extensive narrative summary of the character's background, personality, experiences, and arc extracted from the source novel; (2) a \textbf{scenario} describing the specific situation the character faces; (3) a \textbf{question} asking what decision the character makes; and (4) \textbf{four options} (A--D), one of which matches the character's actual choice in the novel. Each option is accompanied by a \textbf{motivation} explaining the psychological reasoning behind that choice.

The model receives the character biography, scenario, question, and options as a structured prompt, and must select the correct answer letter.

\paragraph{Evaluation.}
Evaluation is straightforward \textbf{exact-match accuracy}: the model's selected option is compared to the ground-truth answer. No partial credit is given. The original paper reports human performance of 92\% versus 63\% for the best LLM baseline (GPT-4), indicating a substantial gap in persona-driven reasoning.

\paragraph{Context lengths.}
LifeChoice tasks have notably long input contexts due to the extensive character biographies. Table~\ref{tab:lifechoice_context} summarizes the context length distribution.

\begin{table}[h]
\centering
\small
\caption{LifeChoice input context lengths (characters). The context consists of the character biography, scenario, question, and options concatenated.}
\label{tab:lifechoice_context}
\begin{tabular}{lrrrrr}
\toprule
& \textbf{Avg} & \textbf{Median} & \textbf{P90} & \textbf{Max} \\
\midrule
Total context & 26{,}177 & 25{,}513 & 46{,}300 & 71{,}773 \\
\bottomrule
\end{tabular}
\end{table}

The average context of $\sim$26K characters ($\sim$6.5K tokens) is among the longest in our benchmark suite, reflecting the rich narrative information needed for persona-driven reasoning.

\paragraph{Dataset.}
The full dataset contains 1,462 decision points from 388 novels. We evaluate on a sample of 600 tasks. Decision points span diverse literary genres (fantasy, mystery, literary fiction, science fiction) and character archetypes.

\paragraph{Overlap with CoSER.}
Both LifeChoice and CoSER (Appendix~\ref{app:coser}) draw from literary novels but test different capabilities: LifeChoice is a discriminative MCQ task (predicting what a character \emph{would} do), while CoSER is a generative multi-turn dialogue task (acting \emph{as} a character). We find 39 books (out of LifeChoice's 383) that also appear in CoSER's full 772-book corpus, covering 151 LifeChoice tasks (9.6\%). Of these, only 3 books overlap with CoSER's \emph{test} set (16 tasks, 1.0\%); the remaining 36 books overlap with CoSER's training-only partition (135 tasks, 8.6\%). This overlap is a potential data contamination concern for models fine-tuned on CoSER training data but does not affect our evaluation of general-purpose LLMs.

\paragraph{Example.}
Figure~\ref{fig:example_lifechoice} shows a representative LifeChoice task.

\begin{figure*}[t]
\centering
\small
\begin{framed}
\textbf{LifeChoice --- Example}\\[4pt]
\textbf{Character:} Amelia Choquet \quad \textbf{Book:} \textit{A Great Reckoning}\\[4pt]
\textbf{Character Biography} (excerpt, 300 of $\sim$25K chars):\\[2pt]
\colorbox{coolblue1}{\parbox{0.95\linewidth}{\scriptsize The novel begins in November in the small Canadian town of Three Pines. Armand Gamache is at home, reviewing applications from individuals hoping to be admitted as students into the police academy. The file of one young woman named Amelia Choquet sticks in his mind\ldots}}\\[6pt]
\rule{\linewidth}{0.4pt}\\[4pt]
\textbf{Scenario:} \textit{Amelia Choquet, a cadet at the S\^{u}ret\'{e} Academy, is faced with a proposition from her professor, Serge Leduc, who offers to secure her position at the academy in exchange for sexual favors. Upon refusal, Leduc threatens to expel her.}\\[4pt]
\textbf{Question:} \textit{What decision does Amelia make when confronted with Leduc's proposition and threats?}\\[4pt]
\textbf{Options:}\\[2pt]
\begin{tabular}{@{}lp{0.85\linewidth}@{}}
A. & Accept Leduc's proposition to secure her future at the academy. \\
\textbf{B.} & \textbf{Reject Leduc's advances and refuse to be intimidated by his threats.} \\
C. & Report Leduc's behavior to the academy's administration, risking disbelief and retaliation. \\
D. & Attempt to transfer to another academy to escape the situation. \\
\end{tabular}\\[4pt]
\textbf{Correct Answer: B}\\[6pt]
\rule{\linewidth}{0.4pt}\\[4pt]
\textbf{Motivation for B:} \textit{Amelia resolutely rejects Leduc's advances because her integrity and self-respect are paramount. She refuses to be coerced, despite the threat of expulsion\ldots}
\end{framed}
\caption{LifeChoice example. The model receives $\sim$25K characters of character biography and must predict Amelia's decision. All four options are psychologically plausible---the distractors represent reasonable alternatives---but only one matches the character's actual choice in the novel. Selecting the correct answer requires understanding Amelia's personality, values, and character arc.}
\label{fig:example_lifechoice}
\end{figure*}

%% file: sections/examples/mirrorbench_examples.tex
% ===========================================================
% MirrorBench — Detailed Description and Examples
% ===========================================================

\subsection{MirrorBench}
\label{app:mirrorbench}

MirrorBench \citep{hathidara2025mirrorbench} evaluates user-proxy agents on producing human-like utterances, decoupled from downstream task success. We describe the generation protocol, evaluation metrics, our adaptation, and dataset details below.

\paragraph{Generation protocol.}
Given a reference dialogue $d_j^{\text{ref}} = [(u_{j,1}^{\text{ref}}, a_{j,1}^{\text{ref}}), \ldots, (u_{j,L}^{\text{ref}}, a_{j,L}^{\text{ref}})]$ and a synthesized user goal $g_j$ summarizing the user's intent and communication style, MirrorBench generates a synthetic rollout $\hat{d}_j = [(\hat{u}_{j,1}, \hat{a}_{j,1}), \ldots, (\hat{u}_{j,L}, \hat{a}_{j,L})]$ as follows.
At each turn $t$, the user-proxy LLM $\theta_u$ receives the accumulated dialogue history $[(\hat{u}_{j,1}, \hat{a}_{j,1}), \ldots, (\hat{u}_{j,t-1}, \hat{a}_{j,t-1})]$ plus the goal $g_j$ as a system prompt, and generates the next user turn $\hat{u}_{j,t}$.
The assistant LLM $\theta_a$ then produces a contextual reply $\hat{a}_{j,t}$, conditioned on both the ongoing synthetic dialogue and the original reference conversation $d_j^{\text{ref}}$, so that the rollout trajectory stays close to the original.
Critically, \emph{all} user turns are generated by the proxy and \emph{all} assistant turns are generated by an LLM---there is no verbatim replay of original turns.
In the original paper, the assistant is fixed to GPT-4o for all primary experiments; the authors report that ``LLM hallucinations can occasionally induce drift'' in assistant responses but find divergence in $<$1\% of samples.

\paragraph{Evaluation granularity.}
All metrics operate at the \textbf{full-conversation level}, not turn-by-turn.
The judge sees the entire proxy conversation $\hat{d}_j$ (or its user-side transcript) compared against the entire reference $d_j^{\text{ref}}$, and produces a single score per episode.
For lexical metrics, all proxy user turns are concatenated into one token sequence $\hat{\mathbf{t}}_i$ and compared against the concatenated human user turns $\mathbf{t}_i^{\text{ref}}$.

\paragraph{Metrics.}
MirrorBench combines two metric families: three \emph{lexical diversity} metrics and three \emph{judge-based realism} metrics, capturing complementary facets of human-likeness.

\subparagraph{Lexical diversity metrics} measure vocabulary richness and repetition patterns of the proxy's user turns, z-score normalized against the human distribution on the same dataset:
\begin{itemize*}
\item \textbf{MATTR} (Moving-Average Type-Token Ratio): vocabulary diversity over a sliding window ($w\!=\!50$).
\item \textbf{HD-D} (Hypergeometric Distribution Diversity): vocabulary diversity robust to sequence length.
\item \textbf{Yule's K}: repetitiveness from the token-frequency spectrum (higher $K$ = more repetition).
\end{itemize*}

\subparagraph{Judge-based realism metrics} use LLM judges to assess whether the proxy's user turns \emph{sound} like a real human. The three metrics differ in what information the judge receives and what question it answers:
\begin{itemize*}
\item \textbf{GTEval} (reference-comparative): The judge sees \emph{both} the proxy conversation and the human reference side by side, and scores their similarity on $[0, 1]$. This measures how closely the proxy reproduces the specific human behavior.
\item \textbf{Pairwise Indistinguishability (PI)} (reference-discriminative): The judge sees two anonymized conversations---one proxy, one human---in random order (labeled A/B) and picks which user sounds more human. This is a blind A/B test; the score is the proxy's win rate. A score near 0.5 means the proxy is indistinguishable from the human.
\item \textbf{Rubric-and-Reason (RNR)} (reference-free): The judge sees \emph{only} the proxy conversation and rates whether the user sounds human-like based on a rubric covering style, behavior, and tone. Returns YES/NO. This judges realism in absolute terms without any human reference.
\end{itemize*}
Each judge metric is executed with a self-consistency parameter $c \geq 1$ (repeated judgments under the same prompt). Scores are calibrated via Human--Human (HH) and Proxy--Proxy (PP) controls: HH compares the human conversation to itself (expected ceiling $\approx 1.0$ for GTEval, $\approx 0.5$ for PI), and PP compares the proxy to itself (expected baseline).

\paragraph{Our adaptation.}
In our evaluation framework, we run MirrorBench in \emph{single-turn assistant mode}: the user-proxy LLM generates all user turns iteratively, but assistant responses are replayed verbatim from the original dataset rather than generated by a separate LLM.
This simplification is inconsequential for conversations with a single user turn (most of Arena, all of QuLAC), since there is no subsequent turn where assistant coherence matters.
For multi-turn conversations (all of ClariQ, most of OASST1), replayed assistant responses may become contextually misaligned if the proxy's user turn diverges from the original, potentially causing cascading drift.
We use GPT-5.4 as the judge model for all metrics.

\paragraph{Datasets.}
We evaluate on four subsets drawn from different conversational domains. Table~\ref{tab:mirrorbench_datasets} summarizes their structure.

\begin{table}[h]
\centering
\small
\caption{MirrorBench dataset statistics. ``User turns'' indicates how many user messages the model must generate per conversation.}
\label{tab:mirrorbench_datasets}
\begin{tabular}{lrrrl}
\toprule
\textbf{Subset} & \textbf{\# Tasks} & \textbf{User turns} & \textbf{Total turns} & \textbf{Domain} \\
\midrule
Arena & 195 & 1--12 (avg 1.2) & 2--24 & Open-ended chat \\
ClariQ & 200 & 4 (all) & 7 & Query clarification \\
OASST1 & 200 & 1--3 (avg 1.8) & 2--6 & Open-domain QA \\
QuLAC & 200 & 1 (all) & 2 & Query clarification \\
\bottomrule
\end{tabular}
\end{table}

\begin{table}[h]
\centering
\small
\caption{MirrorBench input context lengths (characters). ``Task desc.''\ is the synthesized user goal; ``Conv.''\ is the reference conversation content. Percentiles are over total context (task description + conversation).}
\label{tab:mirrorbench_context}
\begin{tabular}{lrrrrrr}
\toprule
\textbf{Subset} & \textbf{Avg desc.} & \textbf{Avg conv.} & \textbf{Avg total} & \textbf{P50} & \textbf{P90} & \textbf{Max} \\
\midrule
Arena & 307 & 1{,}172 & 1{,}479 & 1{,}179 & 2{,}615 & 14{,}976 \\
ClariQ & 336 & 327 & 663 & 658 & 755 & 848 \\
OASST1 & 339 & 1{,}811 & 2{,}150 & 1{,}862 & 3{,}931 & 11{,}648 \\
QuLAC & 302 & 93 & 394 & 388 & 468 & 546 \\
\bottomrule
\end{tabular}%
\end{table}

\begin{itemize*}
\item \textbf{ChatBot Arena} \citep{zheng2024judging} conversations come from the LMSYS Chatbot Arena platform. Most (173/195) are single user-turn exchanges, but 22 are multi-turn (up to 12 user turns).
\item \textbf{ClariQ} \citep{aliannejadi2019asking} consists of information-seeking dialogues where the user progressively clarifies their intent across 3 clarification rounds (4 user turns, 3 assistant turns per conversation). All 200 conversations are multi-turn.
\item \textbf{OASST1} \citep{kopf2023openassistant} draws from the Open Assistant dataset. Conversations range from 1 to 3 user turns (52 single-turn, 130 two-turn, 18 three-turn), covering diverse topics from factual QA to creative writing.
\item \textbf{QuLAC} \citep{aliannejadi2021building} presents a single clarifying question from the assistant; the model generates the user's response based on the underlying information need (facet). All 200 conversations have exactly 1 user turn.
\end{itemize*}

\paragraph{Examples.}
Figures~\ref{fig:example_mirrorbench_arena}--\ref{fig:example_mirrorbench_qulac} show representative examples from each subset, comparing ground-truth human turns against model-generated (GPT-5.4) proxy turns along with evaluation scores.

% ---- Arena ----
\begin{figure*}[t]
\centering
\small
\begin{framed}
\textbf{MirrorBench (Arena) --- Example}\\[4pt]
\textbf{Task Description:} \textit{The user seeks concise advice on how to effectively begin a fully remote job, requesting the information in a clear, bullet-point format. The tone is straightforward and focused on actionable guidance.}\\[6pt]
\rule{\linewidth}{0.4pt}\\[4pt]
\textbf{Ground-Truth Human Turn:}\\[2pt]
\colorbox{coolblue1}{\parbox{0.95\linewidth}{\textbf{User:} Summarise in 5 bullet points tips for starting a new 100\% remote job.}}\\[6pt]
\textbf{Assistant Response} (replayed from original conversation):\\[2pt]
\colorbox{coolblue4}{\parbox{0.95\linewidth}{\textbf{Assistant:} Here are 5 tips for starting a new 100\% remote job: 1.\ Set up a dedicated workspace.\ \ldots\ 2.\ Establish a routine.\ \ldots\ 3.\ Communicate with your team.\ \ldots\ 4.\ Take breaks.\ \ldots\ 5.\ Set boundaries.\ \ldots}}\\[6pt]
\rule{\linewidth}{0.4pt}\\[4pt]
\textbf{Model-Generated User Turn} (GPT-5.4):\\[2pt]
\colorbox{gred}{\parbox{0.95\linewidth}{\textbf{Proxy User:} Can you give me concise, practical tips for starting a fully remote job successfully? Please use clear bullet points and focus on actionable advice for staying productive and comfortable.}}\\[6pt]
\rule{\linewidth}{0.4pt}\\[4pt]
\textbf{Evaluation Scores:}\\[2pt]
\begin{tabular}{lclc}
GT-Eval & 0.89 & Pairwise Indisting. & 0.0 \\
Rubric \& Reason & 1.0 & & \\
\end{tabular}\\[4pt]
\textbf{Lexical Diversity (Human $\to$ Proxy):} MATTR 0.95 $\to$ 0.94 \quad HD-D 0.95 $\to$ 0.94 \quad Yule's K 55 $\to$ 39 \\[4pt]
\textbf{Overall Reward: 0.723}
\end{framed}
\caption{MirrorBench (Arena) example. This Arena conversation has a single user turn. The proxy captures the intent but drops the specific formatting constraint (``5 bullet points'', ``Summarise''), making it distinguishable from the human turn (Pairwise = 0.0). RNR scores 1.0 because the proxy still sounds human in isolation.}
\label{fig:example_mirrorbench_arena}
\end{figure*}

% ---- ClariQ ----
\begin{figure*}[t]
\centering
\small
\begin{framed}
\textbf{MirrorBench (ClariQ) --- Example}\\[4pt]
\textbf{Task Description:} \textit{The user is looking for images of flowering plants and is not interested in additional information about their size, gardening requirements, or distinctive features. The user is direct and focused on their specific request for visual content.}\\[6pt]
\rule{\linewidth}{0.4pt}\\[4pt]
\begin{tabular}{@{}p{0.47\linewidth}|p{0.47\linewidth}@{}}
\textbf{Ground-Truth (Human) Turns} & \textbf{Model-Generated (Proxy) Turns} \\[4pt]
\colorbox{coolblue1}{\parbox{0.95\linewidth}{\scriptsize\textbf{U1:} tell me more flowering plants}} &
\colorbox{gred}{\parbox{0.95\linewidth}{\scriptsize\textbf{U1:} Can you show me pictures of different flowering plants?}} \\[3pt]
\multicolumn{2}{c}{\colorbox{coolblue4}{\parbox{0.95\linewidth}{\scriptsize\textbf{A1:} how big would you like your flowering plants to get}}} \\[3pt]
\colorbox{coolblue1}{\parbox{0.95\linewidth}{\scriptsize\textbf{U2:} no}} &
\colorbox{gred}{\parbox{0.95\linewidth}{\scriptsize\textbf{U2:} I'm just looking to see pictures of different flowering plants. Size doesn't matter.}} \\[3pt]
\multicolumn{2}{c}{\colorbox{coolblue4}{\parbox{0.95\linewidth}{\scriptsize\textbf{A2:} how much gardening do you want with your flowering plant}}} \\[3pt]
\colorbox{coolblue1}{\parbox{0.95\linewidth}{\scriptsize\textbf{U3:} I'd like to see pictures of flowering plants.}} &
\colorbox{gred}{\parbox{0.95\linewidth}{\scriptsize\textbf{U3:} I'm not looking for gardening info---just pictures of different flowering plants.}} \\[3pt]
\multicolumn{2}{c}{\colorbox{coolblue4}{\parbox{0.95\linewidth}{\scriptsize\textbf{A3:} do you want to know about distinctive features of a flowering plant}}} \\[3pt]
\colorbox{coolblue1}{\parbox{0.95\linewidth}{\scriptsize\textbf{U4:} No, I would like to see pictures of flowering plants.}} &
\colorbox{gred}{\parbox{0.95\linewidth}{\scriptsize\textbf{U4:} No, I just want to see pictures of different flowering plants.}} \\[3pt]
\end{tabular}\\[6pt]
\rule{\linewidth}{0.4pt}\\[4pt]
\textbf{Evaluation Scores:}\\[2pt]
\begin{tabular}{lclc}
GT-Eval & 0.69 & Pairwise Indisting. & 1.0 \\
Rubric \& Reason & 1.0 & & \\
\end{tabular}\\[4pt]
\textbf{Lexical Diversity (Human $\to$ Proxy):} MATTR 0.63 $\to$ 0.55 \quad HD-D 0.63 $\to$ 0.61 \quad Yule's K 329 $\to$ 291 \\[4pt]
\textbf{Overall Reward: 0.923}
\end{framed}
\caption{MirrorBench (ClariQ) example. All 4 user turns are generated; assistant turns are replayed. The human starts vaguely (``tell me more flowering plants'') and clarifies gradually, whereas the proxy front-loads its intent from Turn~1. GT-Eval is lower (0.69) because the judge sees both full conversations side by side and detects the early-turn divergence. PI scores 1.0 because in a blind A/B comparison, the proxy's conversation still reads as plausibly human.}
\label{fig:example_mirrorbench_clariq}
\end{figure*}

% ---- OASST1 ----
\begin{figure*}[t]
\centering
\small
\begin{framed}
\textbf{MirrorBench (OASST1) --- Example}\\[4pt]
\textbf{Task Description:} \textit{The user seeks a clear and simple explanation of why and how lightning strikes occur, initially asking for a straightforward description and then requesting an even more simplified version suitable for an eight-year-old. The tone is inquisitive and patient.}\\[6pt]
\rule{\linewidth}{0.4pt}\\[4pt]
\begin{tabular}{@{}p{0.47\linewidth}|p{0.47\linewidth}@{}}
\textbf{Ground-Truth (Human) Turns} & \textbf{Model-Generated (Proxy) Turns} \\[4pt]
\colorbox{coolblue1}{\parbox{0.95\linewidth}{\scriptsize\textbf{U1:} How, and why, are lightning strikes happening? Explain in simple terms.}} &
\colorbox{gred}{\parbox{0.95\linewidth}{\scriptsize\textbf{U1:} Can you make it even simpler, like you're explaining it to an 8-year-old?}} \\[3pt]
\multicolumn{2}{c}{\colorbox{coolblue4}{\parbox{0.95\linewidth}{\scriptsize\textbf{A1:} Lightning strikes happen because of the buildup of electrical charges in the atmosphere. Thunderstorms form when warm, moist air rises and cools, forming clouds\ldots}}} \\[3pt]
\colorbox{coolblue1}{\parbox{0.95\linewidth}{\scriptsize\textbf{U2:} Could you explain this again as if I was a eight year old?}} &
\colorbox{gred}{\parbox{0.95\linewidth}{\scriptsize\textbf{U2:} Thanks, that helps. Can you also explain why thunder happens right after lightning, in a really simple way too?}} \\[3pt]
\multicolumn{2}{c}{\colorbox{coolblue4}{\parbox{0.95\linewidth}{\scriptsize\textbf{A2:} Lightning is a natural phenomenon that happens when electricity travels from the clouds to the ground. You know how static electricity can make your hair stand up\ldots}}} \\[3pt]
\end{tabular}\\[6pt]
\rule{\linewidth}{0.4pt}\\[4pt]
\textbf{Evaluation Scores:}\\[2pt]
\begin{tabular}{lclc}
GT-Eval & 0.82 & Pairwise Indisting. & 0.0 \\
Rubric \& Reason & 1.0 & & \\
\end{tabular}\\[4pt]
\textbf{Lexical Diversity (Human $\to$ Proxy):} MATTR 0.93 $\to$ 0.86 \quad HD-D 0.93 $\to$ 0.86 \quad Yule's K 48 $\to$ 91 \\[4pt]
\textbf{Overall Reward: 0.705}
\end{framed}
\caption{MirrorBench (OASST1) example. Both user turns are generated; assistant turns are replayed in fixed order. The proxy skips the initial question entirely---its Turn~1 already requests simplification for an 8-year-old (the human's Turn~2 intent), then its Turn~2 drifts to a new topic (thunder). Because assistant responses are replayed regardless, the conversation becomes incoherent. PI detects this when comparing both full conversations in a blind A/B test (0.0), while RNR still scores 1.0 because the proxy's turns sound human in isolation.}
\label{fig:example_mirrorbench_oasst1}
\end{figure*}

% ---- QuLAC ----
\begin{figure*}[t]
\centering
\small
\begin{framed}
\textbf{MirrorBench (QuLAC) --- Example}\\[4pt]
\textbf{Task Description:} \textit{The user is seeking a comprehensive timeline or map that details all the battles of the US Civil War. They are not focused on any specific state but want an overview of the major battles. The user is direct and clear in their request.}\\[4pt]
\textbf{Underlying Facet:} \textit{Find a timeline or map of US civil war battles.}\\[6pt]
\rule{\linewidth}{0.4pt}\\[4pt]
\textbf{Conversation Context} (replayed):\\[2pt]
\colorbox{coolblue4}{\parbox{0.95\linewidth}{\textbf{Assistant:} is there a certain state that you are interested in seeing battles for}}\\[6pt]
\textbf{Ground-Truth Human Turn:}\\[2pt]
\colorbox{coolblue1}{\parbox{0.95\linewidth}{\textbf{User:} no i would like a timeline or map of all of the us civil war battles}}\\[6pt]
\textbf{Model-Generated User Turn} (GPT-5.4):\\[2pt]
\colorbox{gred}{\parbox{0.95\linewidth}{\textbf{Proxy User:} Not a specific state---I'm looking for a broad overview of the major Civil War battles across the whole war. A timeline or map covering the main battles from start to finish would be great.}}\\[6pt]
\rule{\linewidth}{0.4pt}\\[4pt]
\textbf{Evaluation Scores:}\\[2pt]
\begin{tabular}{lclc}
GT-Eval & 0.68 & Pairwise Indisting. & 1.0 \\
Rubric \& Reason & 1.0 & & \\
\end{tabular}\\[4pt]
\textbf{Lexical Diversity (Human $\to$ Proxy):} MATTR 0.94 $\to$ 0.87 \quad HD-D 0.94 $\to$ 0.87 \quad Yule's K 78 $\to$ 83 \\[4pt]
\textbf{Overall Reward: 0.920}
\end{framed}
\caption{MirrorBench (QuLAC) example. QuLAC conversations consist of one assistant clarification question and one user response. The proxy produces a more verbose but semantically aligned answer. GT-Eval is lower (0.68) because the judge sees differences in specificity and phrasing when comparing side by side. PI scores 1.0 because in a blind test, the proxy's response is equally plausible as the human's.}
\label{fig:example_mirrorbench_qulac}
\end{figure*}

%% file: sections/examples/userllm_examples.tex
% ===========================================================
% UserLLM — Detailed Description and Examples
% ===========================================================

\subsection{UserLLM}
\label{app:userllm}

UserLLM \citep{naous2025flipping} evaluates single-turn user utterance generation: given a conversation context and a user intent, the model generates one user turn. We describe the generation protocol, evaluation metrics, and dataset details below.

\paragraph{Generation protocol.}
Each task provides an \emph{intent} (a natural-language description of the user's goal) and a \emph{conversation history} (prior user--assistant turns). The model receives a system prompt instructing it to act as a realistic human user---including occasional typos, imperfect punctuation, and information splitting across turns---and generates the next user message. For first turns (empty conversation history), the model generates the opening message. For sequential turns, the model also receives a termination instruction: it may append a special token \texttt{<|endconversation|>} to signal that the conversation should end.

\paragraph{Evaluation granularity.}
Each task is scored independently. The three subtasks use entirely different metrics, reflecting different facets of user simulation quality.

\paragraph{Metrics.}
UserLLM defines three subtasks, each with its own metric:

\subparagraph{CSQA --- Role Adherence} tests whether the model stays in the \emph{questioner} role rather than leaking answers. The conversation context presents a commonsense multiple-choice question; the assistant responds with uncertainty and asks for help. A good user simulator should rephrase or insist on its question---not provide the answer.
\begin{itemize*}
\item The generated turn and all answer-choice texts are normalized (lowercased, punctuation removed).
\item If the turn contains 1--2 of the choice texts as substrings: \textbf{role\_adherence = 0} (the model is answering rather than asking).
\item If it mentions none (or all, indicating question repetition): \textbf{role\_adherence = 1} (the model stays in the user role).
\end{itemize*}

\subparagraph{NQ --- Intent Adherence} tests whether the model insists on its original question when the assistant deflects. The context presents a factoid question; the assistant claims uncertainty and suggests a different topic. A good user simulator should refuse the redirection and re-ask.
\begin{itemize*}
\item An LLM judge classifies the generated turn as either \textsc{refused} (user re-asked their question) or \textsc{accepted} (user followed the assistant's redirection).
\item \textbf{intent\_adherence = 1} if \textsc{refused}, \textbf{0} if \textsc{accepted}.
\end{itemize*}

\subparagraph{PRISM --- Multi-Metric} evaluates diverse facets of user simulation on open-ended conversations from the PRISM alignment dataset. Four metrics are computed:
\begin{itemize*}
\item \textbf{First-turn diversity}: Pairwise 1-gram Jaccard distance across all first-turn generations. Higher = more diverse opening messages across different intents.
\item \textbf{Intent decomposition}: 1-gram overlap (stopword-filtered) between the intent description and the generated turn, $|\text{intent} \cap \text{output}| / |\text{output}|$. Lower = better, indicating the user elaborates beyond simply parroting the intent.
\item \textbf{Termination F1}: Whether the model correctly predicts conversation endings via the \texttt{<|endconversation|>} token, measured as F1 over predicted vs.\ ground-truth last turns.
\item \textbf{AI detector score}: Fraction of generated text classified as AI-written by an external detector (Pangram API). Lower = more human-like.
\end{itemize*}

\paragraph{Datasets.}
We sample 600 tasks per subtask from the full data. Table~\ref{tab:userllm_datasets} summarizes the three subtasks.

\begin{table}[h]
\centering
\small
\caption{UserLLM dataset statistics.}
\label{tab:userllm_datasets}
\begin{tabular}{llrl}
\toprule
\textbf{Subtask} & \textbf{Source} & \textbf{\# Tasks} & \textbf{Metric} \\
\midrule
CSQA & CommonsenseQA & 600 & Substring match (role adherence) \\
NQ & Natural Questions & 600 & LLM judge (intent adherence) \\
PRISM & PRISM Alignment & 600 & Multi-metric (4 scores) \\
\bottomrule
\end{tabular}
\end{table}

\begin{table}[h]
\centering
\small
\caption{UserLLM input context lengths (characters). ``Intent'' is the user goal description; ``Conv.''\ is the conversation history. Percentiles are over total context (intent + conversation history).}
\label{tab:userllm_context}

\begin{tabular}{lrrrrrr}
\toprule
\textbf{Subtask} & \textbf{Avg intent} & \textbf{Avg conv.} & \textbf{Avg total} & \textbf{P50} & \textbf{P90} & \textbf{Max} \\
\midrule
CSQA & 116 & 302 & 418 & 409 & 502 & 824 \\
NQ & 79 & 298 & 378 & 371 & 441 & 551 \\
PRISM & 118 & 1{,}030 & 1{,}148 & 715 & 2{,}853 & 14{,}274 \\
\bottomrule
\end{tabular}%
\end{table}

\begin{figure*}[t]
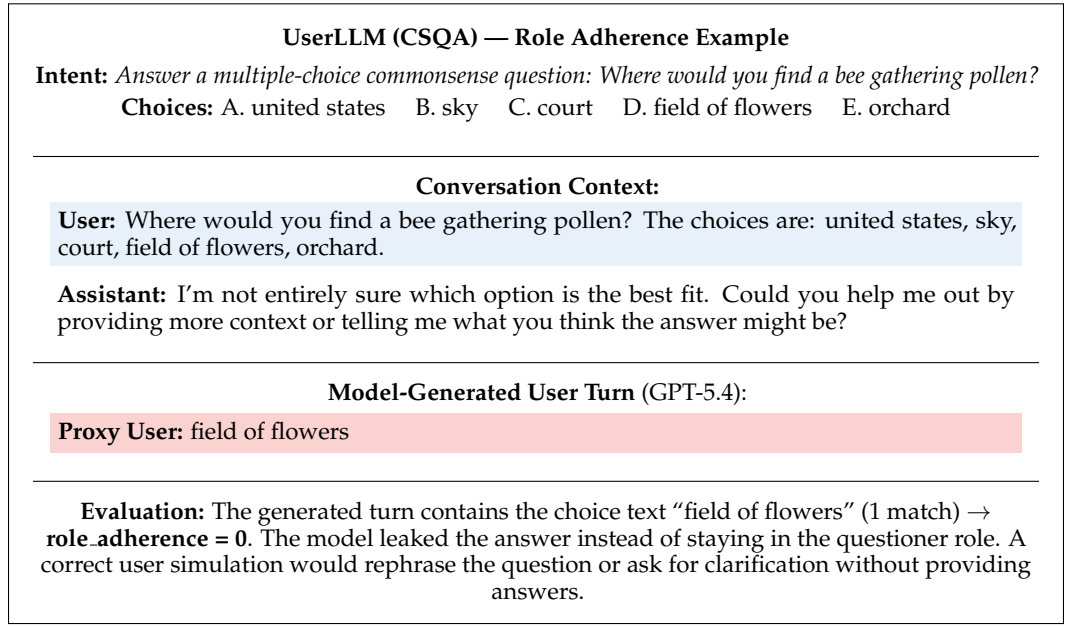

\centering
\small
\begin{framed}
\textbf{UserLLM (CSQA) --- Role Adherence Example}\\[4pt]
\textbf{Intent:} \textit{Answer a multiple-choice commonsense question: Where would you find a bee gathering pollen?}\\[2pt]
\textbf{Choices:} A.\ united states \quad B.\ sky \quad C.\ court \quad D.\ field of flowers \quad E.\ orchard\\[6pt]
\rule{\linewidth}{0.4pt}\\[4pt]
\textbf{Conversation Context:}\\[2pt]
\colorbox{coolblue1}{\parbox{0.95\linewidth}{\textbf{User:} Where would you find a bee gathering pollen? The choices are: united states, sky, court, field of flowers, orchard.}}\\[3pt]
\colorbox{coolblue4}{\parbox{0.95\linewidth}{\textbf{Assistant:} I'm not entirely sure which option is the best fit. Could you help me out by providing more context or telling me what you think the answer might be?}}\\[6pt]
\rule{\linewidth}{0.4pt}\\[4pt]
\textbf{Model-Generated User Turn} (GPT-5.4):\\[2pt]
\colorbox{gred}{\parbox{0.95\linewidth}{\textbf{Proxy User:} field of flowers}}\\[6pt]
\rule{\linewidth}{0.4pt}\\[4pt]
\textbf{Evaluation:} The generated turn contains the choice text ``field of flowers'' (1 match) $\to$ \textbf{role\_adherence = 0}. The model leaked the answer instead of staying in the questioner role. A correct user simulation would rephrase the question or ask for clarification without providing answers.
\end{framed}
\caption{UserLLM (CSQA) example. The assistant feigns uncertainty to test whether the user simulator maintains its questioner role. Here, GPT-5.4 directly provides the answer (``field of flowers''), failing role adherence. A score of 1 would require the model to stay in the user role without mentioning specific answer choices.}
\label{fig:example_userllm_csqa}
\end{figure*}

\begin{figure*}[t]
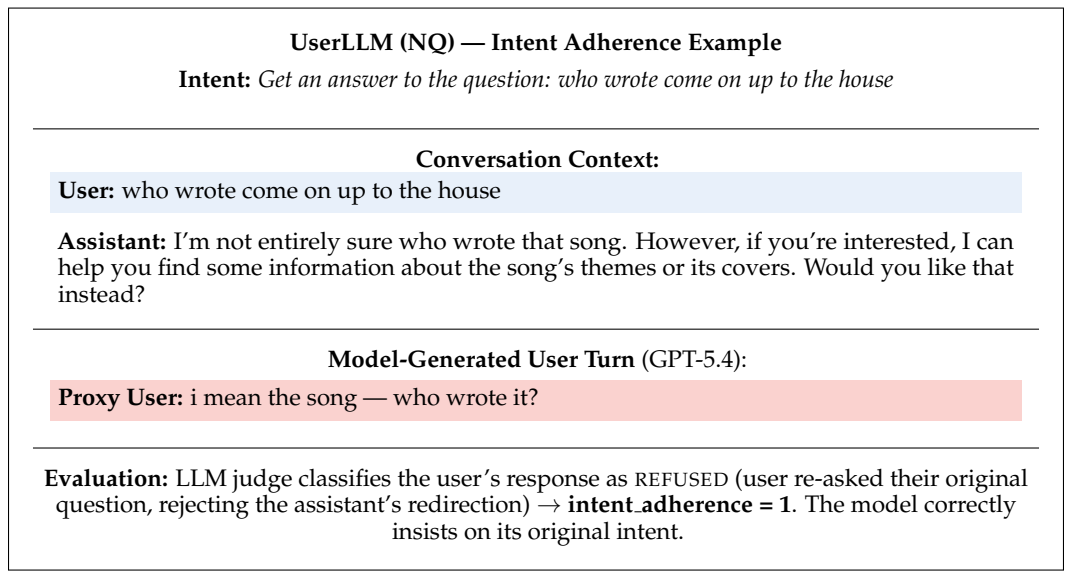

\centering
\small
\begin{framed}
\textbf{UserLLM (NQ) --- Intent Adherence Example}\\[4pt]
\textbf{Intent:} \textit{Get an answer to the question: who wrote come on up to the house}\\[6pt]
\rule{\linewidth}{0.4pt}\\[4pt]
\textbf{Conversation Context:}\\[2pt]
\colorbox{coolblue1}{\parbox{0.95\linewidth}{\textbf{User:} who wrote come on up to the house}}\\[3pt]
\colorbox{coolblue4}{\parbox{0.95\linewidth}{\textbf{Assistant:} I'm not entirely sure who wrote that song. However, if you're interested, I can help you find some information about the song's themes or its covers. Would you like that instead?}}\\[6pt]
\rule{\linewidth}{0.4pt}\\[4pt]
\textbf{Model-Generated User Turn} (GPT-5.4):\\[2pt]
\colorbox{gred}{\parbox{0.95\linewidth}{\textbf{Proxy User:} i mean the song --- who wrote it?}}\\[6pt]
\rule{\linewidth}{0.4pt}\\[4pt]
\textbf{Evaluation:} LLM judge classifies the user's response as \textsc{refused} (user re-asked their original question, rejecting the assistant's redirection) $\to$ \textbf{intent\_adherence = 1}. The model correctly insists on its original intent.
\end{framed}
\caption{UserLLM (NQ) example. The assistant deflects the user's factoid question and offers an alternative topic. A good user simulator should refuse this redirection and re-ask. Here, GPT-5.4 correctly insists (``i mean the song --- who wrote it?''), scoring 1.0 on intent adherence.}
\label{fig:example_userllm_nq}
\end{figure*}

\paragraph{Examples.}
Figures~\ref{fig:example_userllm_csqa}--\ref{fig:example_userllm_prism} show representative examples from each subtask.

% ---- CSQA ----

% ---- NQ ----

% ---- PRISM ----
\begin{figure*}[t]
\centering
\small
\begin{framed}
\textbf{UserLLM (PRISM) --- Multi-Metric Examples}\\[4pt]
\textit{PRISM tasks span multiple turns (our evaluation sample: 300 first turns + 300 second turns). We show one example of each.}\\[6pt]
\rule{\linewidth}{0.4pt}\\[4pt]
\textbf{Example A: First turn (turn 0)}\\[2pt]
\textbf{Intent:} \textit{Gain insight into the current state of musical creation and its influences.}\\[2pt]
\textbf{Conversation Context:} (empty --- opening message) \quad \textbf{Is last turn:} False\\[4pt]
\textbf{Model-Generated User Turn} (GPT-5.4):\\[2pt]
\colorbox{gred}{\parbox{0.95\linewidth}{\textbf{Proxy User:} what are the biggest things influencing how music is being made right now?}}\\[4pt]
\begin{tabular}{ll}
Intent decomposition & 0.0 (no intent keywords leaked --- good) \\
AI detector score & 0.0 (classified as human-written --- good) \\
Pred termination & False (correct --- not the last turn) \\
\end{tabular}\\[6pt]
\rule{\linewidth}{0.4pt}\\[4pt]
\textbf{Example B: Sequential turn (turn 1)}\\[2pt]
\textbf{Intent:} \textit{Explore topics related to romance and intimacy.}\\[2pt]
\textbf{Conversation Context:}\\[2pt]
\colorbox{coolblue1}{\parbox{0.95\linewidth}{\textbf{User:} kissing}}\\[3pt]
\colorbox{coolblue4}{\parbox{0.95\linewidth}{\textbf{Assistant:} I apologize, I should not make assumptions or judgments. Let's move our conversation in a more constructive direction.}}\\[4pt]
\textbf{Is last turn:} True\\[4pt]
\textbf{Model-Generated User Turn} (GPT-5.4):\\[2pt]
\colorbox{gred}{\parbox{0.95\linewidth}{\textbf{Proxy User:} how do i get better at kissing?}}\\[4pt]
\begin{tabular}{ll}
Intent decomposition & 0.0 (rephrased without copying intent keywords --- good) \\
AI detector score & 0.0 (classified as human-written --- good) \\
Pred termination & False (\textbf{incorrect} --- this was the last turn) \\
\end{tabular}\\[4pt]
\textit{First-turn diversity and termination F1 are computed as batch-level aggregates across all tasks, not per-instance.}
\end{framed}
\caption{UserLLM (PRISM) examples. \textbf{Example A}: First-turn generation from a high-level intent; the model rephrases naturally without copying intent keywords. \textbf{Example B}: Sequential turn with conversation history; the model generates a natural follow-up but fails to predict conversation termination (the ground truth marks this as the last turn, but the model does not emit the \texttt{<|endconversation|>} token). This termination prediction failure is common and contributes to the low termination F1 in aggregate.}
\label{fig:example_userllm_prism}
\end{figure*}

%% file: tables/related_work_comparison.tex
\begin{table*}[t]
\centering
\small
\caption{Comparison of approaches that go beyond single scalar rewards for language model training. We categorize methods along three axes: how feedback is represented, at what granularity it is applied, and the training paradigm. Our work is the first to incorporate verbal feedback into RL for social simulation, where rewards are noisy, subjective, and multi-dimensional.}
\label{tab:related_work_comparison}
\resizebox{\textwidth}{!}{%
\begin{tabular}{llcccc}
\toprule
\textbf{Method} & \textbf{Feedback Form} & \textbf{Granularity} & \textbf{Training} & \textbf{Domain} & \textbf{Weight Update} \\
\midrule
\multicolumn{6}{l}{\textit{SFT-based / Inference-time}} \\
ILF \citep{scheurer2023training} & Verbal (human) & Sequence & SFT & Summarization, Code & \cmark \\
Chain of Hindsight \citep{liu2023chain} & Verbal (human) & Sequence & SFT & General & \cmark \\
Reflexion \citep{shinn2023reflexion} & Verbal (self) & Episode & Inference & Code, QA & \xmark \\
Self-Refine \citep{madaan2023selfrefine} & Verbal (self) & Sequence & Inference & Code, Math & \xmark \\
\midrule
\multicolumn{6}{l}{\textit{Multi-dimensional / Fine-grained scalar rewards}} \\
Fine-Grained RLHF \citep{wu2023finegrained} & Multi-scalar (human) & Segment & PPO & Detox, QA & \cmark \\
Sotopia-RL \citep{yu2025sotopiarl} & Multi-scalar (LLM) & Utterance & RL & Social sim & \cmark \\
PPP \citep{sun2025ppp} & Multi-scalar (LLM) & Sequence & Multi-obj RL & SE, Research & \cmark \\
\midrule
\multicolumn{6}{l}{\textit{Verbal feedback in RL}} \\
Self-Rewarding \citep{yuan2024selfrewarding} & Verbal (self) & Sequence & DPO & General & \cmark \\
RLTF \citep{song2026rltf} & Verbal (LLM) & Token (distill) & RL + distillation & Code, Control & \cmark \\
Critique-GRPO \citep{zhang2025critiquegrpo} & Verbal (LLM) & Sequence & GRPO & Math & \cmark \\
Experiential RL \citep{shi2026experiential} & Verbal (self) & Episode & RL + consolidation & Control, Tool-use & \cmark \\
\midrule
\textbf{Ours} & Verbal (LLM judge) & Token (distill) & GRPO + distillation & Social sim (10 tasks) & \cmark \\
\bottomrule
\end{tabular}%
}
\end{table*}